\documentclass[11pt]{article}
\usepackage[utf8]{inputenc}
\usepackage[american]{babel}
\usepackage[T1]{fontenc}
\usepackage{newpxtext}
\usepackage{graphics}
\usepackage{grffile}
\usepackage{fancyhdr}
\usepackage{float}
\usepackage{amsmath}
\usepackage{amssymb}
\usepackage{amsfonts}
\usepackage[font=small,labelfont=bf]{caption}
\usepackage[export]{adjustbox}
\usepackage{mathrsfs}
\usepackage[table,xcdraw]{xcolor}
\usepackage{lipsum}
\usepackage{caption}
\usepackage{booktabs}
\usepackage{subcaption}
\usepackage{latexsym}
\usepackage{amsthm} 
\usepackage{eucal} 
\usepackage[hang]{footmisc}
\usepackage{lmodern}
\usepackage{pstricks}
\usepackage{newlfont}
\usepackage[margin = 0.8in]{geometry}
\usepackage{enumitem}
\usepackage{hyperref}
\renewenvironment{abstract}
 {\small
  \begin{center}
  \bfseries \Large\abstractname\vspace{-.5em}\vspace{0pt}
  \end{center}
  \list{}{
    \setlength{\leftmargin}{0.8cm}%
    \setlength{\rightmargin}{\leftmargin}%
  }%
  \item\relax}
 {\endlist}

\theoremstyle{definition}

\theoremstyle{remark}
\newtheorem{remark}{Remark}[section]
 
\providecommand{\keywords}[1]
{
  \textbf{\textit{Keywords ---}} #1
}
\usepackage{titling}
\title{A Neural Network Ensemble Approach to System Identification}
\author{{Elisa Negrini} \thanks{Mathematical Sciences Department, Worcester Polytechnic Institute, 100 Institute Road, Worcerster, MA, 01609, USA} \and Giovanna Citti \thanks{Department of Mathematics, University of Bologna, Piazza di Porta S. Donato 5, 40126 Bologna BO, Italy} \and Luca Capogna \thanks{Department of Mathematics and
Statistics, Smith College, 7 College Lane, Northampton, MA, 01063, USA}}
\thanksmarkseries{arabic}
\date{\vspace{-5ex}}

\begin{document}
\maketitle
\let\thefootnote\relax\footnote{\textbf{Email Addresses:} Elisa Negrini (corresponding author):  \href{mailto:enegrini@wpi.edu}{enegrini@wpi.edu}, \newline \hspace*{8.0em} \textit{permanent address:} 1 Cedar street Worcester MA 01609 USA, \newline \hspace*{8.0em} Giovanna Citti: \href{mailto:giovanna.citti@unibo.it}{giovanna.citti@unibo.it} ,\newline \hspace*{8.0em} Luca Capogna: \href{mailto:lcapogna@smith.edu}{lcapogna@smith.edu} }
\vspace{-1.3cm}
\begin{abstract}
We present a new algorithm for learning unknown governing equations from trajectory data, using and ensemble of neural networks. Given samples of solutions $x(t)$ to an unknown dynamical system $\dot{x}(t)=f(t,x(t))$, we approximate the function $f$ using an ensemble of neural networks. We express the equation in integral form and use Euler method to predict the solution at every successive time step using at each iteration a different neural network as a prior for $f$. This procedure yields M-1 time-independent networks, where M is the number of time steps at which $x(t)$ is observed. Finally, we obtain a single function $f(t,x(t))$ by neural network interpolation. Unlike our earlier work, where we numerically computed the derivatives of data, and used them as target in a Lipschitz regularized neural network to approximate $f$, our new method avoids numerical differentiations, which are unstable in presence of noise.  We test the new algorithm on multiple examples both with and without noise in the data. We empirically show that generalization and recovery of the governing equation improve by adding a Lipschitz regularization term in our loss function and that this method improves our previous one especially in presence of noise, when numerical differentiation provides low quality target data. Finally, we compare our results with the method proposed by Raissi, et al. arXiv:1801.01236 (2018) and with SINDy.

\end{abstract}
\section{Introduction}\label{Intro}
System identification refers to the problem of building mathematical models and approximating governing equations using only observed data from the system. Governing laws and equations have traditionally been derived from expert knowledge and first principles, however in recent years the large amount of data available resulted in a growing interest in data-driven models and approaches for automated dynamical systems discovery. The applications of system identification include any system where the inputs and outputs can be measured, such as industrial processes, control systems, economic data and financial systems, biology and the life sciences, medicine, social systems, and many more (see for instance \cite{billings2013nonlinear} for more examples of applications).

Examples of frequently used approaches for data-driven discovery of nonlinear differential equations are sparse regression, Gaussian processes, applied Koopmanism and dictionary based approaches, among which neural networks. Sparse regression approaches are based on a user-determined library of candidate terms from which the most important ones are selected using sparse regression (see for instance \cite{schaeffer2013sparse}, \cite{brunton2016discovering},  \cite{rudy2017data}, \cite{schaeffer2017learning}). These methods provide interpretable results, but they are usually sensitive to noise and require the user to choose an “appropriate” sets of basis functions. Identification using Gaussian Processes places a Gaussian prior on the unknown coefficients of the differential equation and infers them via maximum likelihood estimation (see for instance \cite{raissi2017machine}, \cite{raissi2018hidden}, \cite{raissi2018numerical}). The Koopman approach is based on the idea that non linear system identification in the state space is equivalent to linear identification of the Koopman operator in the infinite-dimensional space of observables. The power of the Koopman approach is that it allows to study non-linear systems using traditional techniques in numerical linear algebra.
However, since the Koopman operator is infinite-dimensional, in practice one computes a projection of the Koopman operator onto a finite-dimensional subspace of the observables. This approximation may result in models of very high dimension and has proven challenging in practical applications (see for instance \cite{budivsic2012applied}, \cite{nathan2018applied}, \cite{lusch2018deep}). In this work we use a different approach based on neural networks.
Since neural networks are universal approximators, they are a natural choice for nonlinear system identification: depending on the architecture and on the properties of the loss function, they can be used as sparse regression models, they can act as priors on unknown coefficients or completely determine an unknown differential operator (see for instance \cite{narendra1992neural}, \cite{wang2006fully},  \cite{ogunmolu2016nonlinear}, \cite{raissi2018deep}, \cite{berg2019data}, \cite{raissi2018multistep}, \cite{champion2019data}, \cite{qin2019data}, \cite{NEGRINI2021110549}). The common goal among all such methods is learning a nonlinear and potentially
multi-variate mapping $f$, right-hand-side of the differential equation:
\begin{equation}\label{my_eq}
    \dot{x}(t) = f(t,x)
\end{equation}
that can be used to predict the future system states given a set of data describing the present and past states.

Two main approaches can be used to approximate the function $f$ with a neural network. The first approach aims at approximating the function $f$ directly, like we did in our previous paper \cite{NEGRINI2021110549}. In this work, inspired by the work of Oberman and Calder in \cite{oberman2018lipschitz} , we use a Lipschitz regularized neural network to approximate the RHS of the ODE (\ref{my_eq}), directly from observations of the state vector $x(t)$.  The target data for the network is made of discrete approximations of the velocity vector $\dot{x}(t)$, which act as a prior for $f$. To generate the target data we first denoise the trajectory data using spline interpolation, then we approximate the velocity vector using the numerical derivative of the splines. In the rest of the paper we refer to this method as \textit{splines method}. One limitation of this approach is that, in order to obtain accurate approximations of the function $f$, one needs to obtain reliable target data, approximations of the velocity vector, from the observations of $x(t)$. This proved to be hard when a large amount of noise (more that 2\%) was present in the data or when splines could not approximate the trajectories correctly. When instead we could obtain high quality target data, we empirically proved that, thanks to the Lipschitz regularization, our method was robust to noise and able to provide an accurate approximation of the function $f$.\\
The second approach aims at approximating the function $f$ implicitly by expressing the differential equation (\ref{my_eq}) in integral form and enforcing that the network that approximates $f$ satisfies an appropriate update rule. This is the approach used in \cite{raissi2018multistep}, which we refer to as \textit{multistep method}, where the authors train the approximating network to satisfy a linear multistep method. An advantage of this approach over the previous one is that the target data used to train the multistep network is composed only of observations of the state vector $x(t)$. However, noise in the observations of $x(t)$ can still have a strong impact on the quality of the network approximation of $f$. \\
Later on we will compare these methods with our proposed approach.

In this work we build on the second approach and introduce a new idea to overcome the limitations of the methods mentioned above. Similarly to the multistep method, we express the differential equation in integral form and train the network that approximates $f$ to satisfy Euler update rule (with minimal modifications one can use linear multistep methods as well). This implicit approach overcomes the limitations of the splines method, whose results were strongly dependent on the quality of the velocity vector approximations used as target data. Differently than the multistep method, our proposed approach is based on a Lipschitz regularized ensemble of neural networks and it is able to overcome the sensitivity to noise. 
More specifically,  we consider the system of ODEs (\ref{my_eq})
where $x(t) \in \mathbb{R}^d$ is the state vector of a $d$-dimensional dynamical system at time $t \in I \subset \mathbb{R}$, $\dot{x}(t) \in \mathbb{R}^d$ is the first order time derivative of $x(t)$ and $f: \mathbb{R}^{1+d}\rightarrow \mathbb{R}^d$ is a vector-valued function right-hand side of the differential equation. We approximate the unknown function $f$ with an ensemble of neural networks. A neural network ensemble is a learning paradigm where a finite number of networks are jointly used to solve a problem.  An ensemble algorithm is generally constructed in two steps: first multiple component neural networks are trained to produce component predictions; then they are combined to produce a final prediction (for a more precise explanation see \cite{krogh1996learning}). In their work \cite{hansen1990neural}, Hansen and Salamon showed that the generalization ability of a neural network architecture can be significantly improved through ensembling. This is the reason why we use an ensemble of neural networks, instead of only one network as it was done in \cite{raissi2018multistep}.

Our proposed ensemble architecture is composed of two blocks: the first, which we call \textit{target data generator} is an ensemble of neural networks whose goal is to produce accurate velocity vector approximations using only observations of $x(t)$. To train this ensemble of networks, we express equation (\ref{my_eq}) in integral form and use Euler method to predict the solution at every successive time step using at each iteration a different neural network as a prior for $f$. If $M$ denotes the number of time steps at which $x(t)$ is observed, then the procedure described above yields $M-1$ time-independent networks, each of which approximates the velocity vector $\dot{x}(t)$ for a fixed time $t$. The second block of the ensemble architecture is the \textit{interpolation network}. This is a Lipschitz regularized feed forward network $N$ as defined in \cite{NEGRINI2021110549}. This network takes as input an observation of the time $t$ and of the state vector $x(t)$ and uses as target data the approximations of the velocity vector generated by the target data generator. Once trained, the interpolation network provides the desired approximation of the RHS function $f$ on its domain.

Finally we want to comment on our choice of using ensembles of neural networks as compared to the other methods listed above for system identification. In our experience and from a literature review, neural networks are a good choice for function approximation because of their ability to learn and model non-linear and complex functions as well as to generalize to unseen data. For example, it has been shown empirically in \cite{choon2008functional} that neural networks outperform polynomial regression when complicated interactions are present in the function to approximate. We also show in Section \ref{comparisonR} that, for noisy data, our Lipschitz regularized ensemble approach outperforms the splines and mutlistep methods as well as polynomial regression and the dictionary based method SINDy (Sparse Identification of Nonlinear Dynamics) \cite{brunton2016discovering}.\\
Since neural networks are universal approximators, we do not need any prior information about the order or about the analytical form of the differential equation as in \cite{schaeffer2013sparse}, \cite{rudy2017data}, \cite{schaeffer2017learning}, \cite{sahoo2018learning}, \cite{hasan2020learning}; this allows to accurately recover very general and complex RHS functions even when no information on the target function is available.
\\Since the proposed ensemble method is based on weak notion of solution using integration (see formula (\ref{weakEq})) it can be used to reconstruct non-smooth RHS functions (see Example \ref{nsRHS}). This is especially an advantage over models that rely on the notion of classical solution like the Splines Method \cite{NEGRINI2021110549}. The ability of our proposed method to accurately approximate both smooth and non-smooth functions make it an extremely valuable approach when working with real-world data.\\
Another advantage of our ensemble approach is its ability to overcome sensitivity to noise and avoid overfitting. This is due to the fact that we use an ensemble of networks to produce our predictions as well as to the Lipschitz regularization term in the loss function of the interpolation network. The ability of our method to overcome sensitivity to noise is especially an advantage over works that use finite differences and polynomial approximation to extract governing equations from data (\cite{brunton2016discovering}, \cite{rudy2017data}), over the Koopman based methods where noise in the data can impact the quality of the finite dimensional approximation of the Koopman operator (\cite{sinha2019robust}, \cite{haseli2019approximating}), as well as over the multistep method \cite{raissi2018multistep}.\\
Finally, our model is defined componentwise so it can be applied to system of equations of any dimension, making it a valuable approach when dealing with high dimensional real-world data. The flexibility and noise robustness of our approach comes, however, at the cost of loss of interpretability and increased computational cost. Training a neural network ensemble is more computationally expensive than training only one neural network, as it is the case in \cite{raissi2018multistep} and \cite{NEGRINI2021110549}, or than using polynomial regression or SINDy. Moreover, the learned ensemble is usually less interpretable than a sparse model based on a dictionary of elementary functions, especially when the number of network learnable parameters is large. However, the trained ensemble produces very accurate results and implements a function which can be easily used in future computations, for example to generate new trajectories like we do in Section \ref{num_ex}.

The paper is organized as follows: in Section \ref{architecture} we describe the ensemble architecture and the loss function used in the training; in Section \ref{data_eval} we describe how the synthetic data was generated, the metrics used to evaluate our method and we precisely define the generalization gap; in Section \ref{num_ex} we propose numerical examples, we show how the ensemble method is an improvement over our previous method and we compare it with other methods for system identification. In Section \ref{discussion} we discuss our numerical results. Finally, in the conclusion Section we summarize our results and describe possible future directions of research.

\section{The Ensemble Architecture}\label{architecture}
\begin{figure}[H]
    \centering
    \includegraphics[width = 1\linewidth]{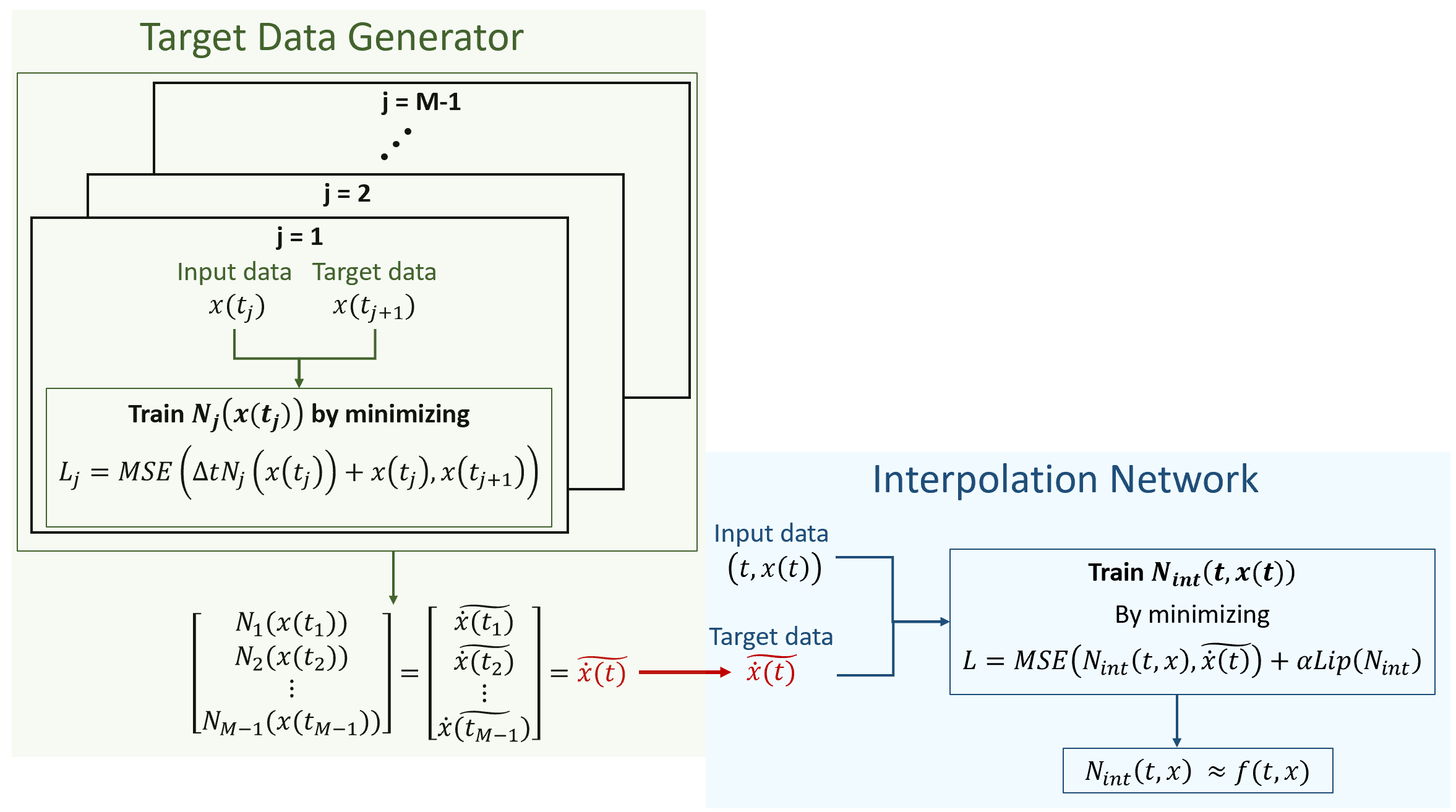}
    \caption{A representation of the ensemble architecture}
    \label{fig:ensemble}
\end{figure}

In this section we describe the architecture used in the experiments.
\medskip\\
In this work, we investigate the problem of approximating unknown governing equations, i.e. approximating the vector-valued RHS $f(t,x)$ of a system of differential equations $\dot{x}(t) = f(t,x)$, directly from discrete observations of the state vector $x(t) \in \mathbb{R}^d$ using an ensemble of feed forward networks, see Figure \ref{fig:ensemble} for a representation of the architecture. \\
We explained before that one limitation of our previously proposed method for system identification (see \cite{NEGRINI2021110549} for the details) is that we used as target data for the network discrete approximations of the velocity vector computed using difference quotients: these provided good approximations of the velocity vector only when small amounts of noise (maximum 2\%) was present in the data. In this work we propose an ensemble approach which is able to provide reliable approximations of the velocity vector from the state vector observations, even when large amounts of noise are present in the data (up to 10\% of noise).

Specifically, the ensemble architecture is composed of two blocks. The first one is the \textit{target data generator}. This is a family of neural networks whose goal is to produce accurate velocity vector approximations using only observations of the state vector $x(t)$. For each time instant $t_j$, we define a neural network $N_j$ which takes as input the state vector at time $t_j$, and it is trained to satisfy Euler update rule to produce an approximation of the state vector at the next time instant. This process implicitly forces the neural network $N_j$ to produce an approximation of the velocity vector at time $t_j$, $\dot{x}(t_j)$. Finally, once all the networks $N_j$ are trained, they collectively provide a discrete approximation of the velocity vector (we use $\widetilde{\quad}$ to indicate an approximation of the quantity under the tilde) :
\begin{equation}
    \begin{bmatrix}
    N_1(x(t_1))\\
    N_2(x(t_2))\\
    \vdots\\
    N_{M-1}(x(t_{M-1}))
    \end{bmatrix} =
     \begin{bmatrix}
     \widetilde{\dot{x}(t_1)}\\
     \widetilde{\dot{x}(t_2)}\\
     \vdots\\
     \widetilde{\dot{x}(t_{M-1})}
     \end{bmatrix} =: \widetilde{\dot{x}(t)}
\end{equation}
The second block, which we call \textit{interpolation network}, is a Lipschitz regularized feed forward network $N_{int}$ as defined in \cite{NEGRINI2021110549}. This network takes as input an observation of the instant time $t$ and of the state vector $x(t)$ and tries to match the target data, $\widetilde{\dot{x}(t)}$, which is made of approximations of the velocity vector generated by the target data generator (first block of the ensemble). Once trained, the interpolation network provides the desired approximation of the RHS function $f$ on its domain:  $N_{int}(t,x) \approx f(t,x)$.

The pipeline for the experiments is as follows: the first step is to train the target data generator to produce reliable velocity vector approximations for the interpolation network. Each network $N_j$ produces an approximation of the velocity vector at time $t_j$, $\dot{x}(t_j)$. These discrete approximations of the velocity vector are then used as target data to train the interpolation network. Once the interpolation network is trained it produces the desired approximation of the function $f(t,x)$.

\subsection{The Target Data Generator}
 The target data generator is a family of neural networks whose goal is to produce reliable velocity vector approximations which will be used as target data for the interpolation network.\\
 The data is selected as follows: given time instants $t_1,\dots, t_M$ and initial conditions $x_1(0), \dots, x_K(0) \in \mathbb{R}^d$, define $$x_i(t_j) \in \mathbb{R}^d, \quad i = 1,\dots,K, \quad j = 1,\dots, M$$ to be an observation of the state vector $x(t)$ at time $t_j$ for initial condition $x_i(0)$.
 
 For each time instant $t_j, \; j = 1, \dots, M-1$ we train a neural network $N_j(x(t_j))$ which approximates the function $f(t,x)$ at time instant $t_j$. More specifically, after training, each neural network $N_j(x(t_j))$ satisfies:
$$ \Delta t \; N_j(x_i(t_j)) + x_i(t_j) \approx x_i(t_{j+1}), \quad \forall i = 1,\dots,K $$
In other words, we express the original ODE $\dot{x} = f(t,x)$ in integral form and use Euler method to predict the solution at every successive time step using, at each iteration, a different neural network as a prior for $f$.

\begin{figure}[H]
    \centering
    \includegraphics[width = 1\linewidth]{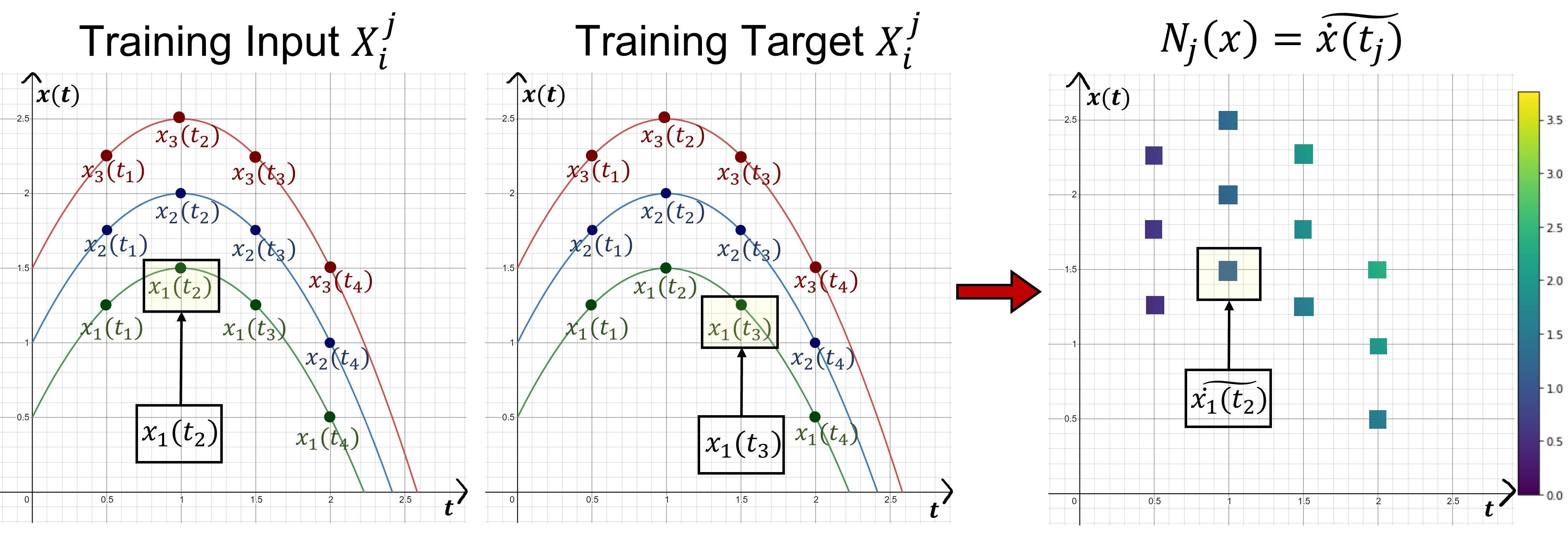}
    \caption{A representation of the data for the target data generator: the inputs are observations of the state vector $x$ for a fixed time $t_j$, $x(t_j)$; the target data are observations of the state vector $x$ at the next time instant $t_{j+1}$, $x(t_{j+1})$. The goal is to train a network $N_j$ which approximates the velocity vector at time $t_j$: this is a prior for the unknown function $f$ at time $t_j$, $f(t_j, x)$.}
    \label{fig:data_gener}
\end{figure}

The data for the target data generator is defined as follows:  for $j = 1, \dots, M-1$ the network data used to train the $j^{th}$ network are couples $(X_i^j, Y_i^j), \; i = 1,\dots, K$, where $X_i^j$ is the  input and $Y_i^j$ is the target and $X_i^j$, $Y_i^j$ are defined as follows:
\begin{align*}
&X_i^j = (x_i(t_j)) \in \mathbb{R}^{d},\\
&Y_i^j = (x_i(t_{j+1})) \in \mathbb{R}^{d}.
\end{align*}
The data is separated into training and testing sets made respectively of 80\% and 20\% of the data. A representation of the data for the interpolation network is provided in Figure \ref{fig:data_gener}.

Each network $N_j$ is a feed forward network with $L_j$ layers and Leaky ReLU activation function. We apply the network to each training input $X_i^j$ and we aim to find the best network parameters to match the corresponding $Y_i^j$.
\smallskip\\
For  $j = 1, \dots, M-1$ and $h = 1,\,2,\,3$ define the weight matrices $W_h^j \in \mathbb{R}^{\;n_h \times n_{h-1}}$ and bias vectors $b_h^j \in \mathbb{R}^{n_h}$ where $n_h \in \mathbb{N}, n_0 = n_3 = d$. Let $\theta^j = \{W^j,b^j\}$ be the model parameters.\\
As activation function, we use a Leaky Rectified Linear Unit (LReLU) with parameter $\varepsilon = 0.01$:
\begin{align*}
    \sigma(x) = \text{LReLU}(x) = \begin{cases}
    \varepsilon x &\text{if } x<0;\\
    x &\text{if } x\geq0.
    \end{cases}
\end{align*}
For an input $X_i^j \in \mathbb{R}^{1+d}$ and parameters $\theta^j$ we have: 
$$N_j(X_i^j, \theta^j) = W_3^j(\sigma ( W_2^j \sigma ( W_1^j X_i^j  + b_1^j) +b_2^j) \dots) + b_3^j \; \in \mathbb{R}^{d}.$$

The loss function $L_j$ used to train each network $N_j$ forces each neural network $N_j$ to satisfy Euler update rule and to produce an approximation of the state vector at the next time instant: 
\begin{equation}\label{weakEq}
    \Delta t \; N_j(x_i(t_j)) + x_i(t_j) \approx x_i(t_{j+1})
\end{equation}

Specifically we define:
\begin{equation*}
    L_j(\theta^j) =\frac{1}{K}\sum_{i=1}^K \|\Delta t \; N_j(x_i(t_j),\theta^j) + x_i(t_j)- x_i(t_{j+1}) \|_2^2 \quad j = 1\dots, M-1
\end{equation*}
Where $\theta^j$ are the network parameters.
The predicted approximation of the function $f(t,x)$ at time $t_j$ is then given by the network $N_j$ corresponding to $\underset{\theta^j}{\mathrm{argmin}}\, L_j(\theta^j)$.

\subsection{The Interpolation Network} 
The interpolation network $N_{int}$ is a Lipschitz regularized neural network which takes as input a time $t$ and an observation of the state vector at time $t$, $x(t)$ and uses as target data the approximation of the velocity vector $\widetilde{\dot{x}(t)}$ given by the target data generator (this acts as a prior for the unknown function $f(t,x)$).  Once trained the interpolation network $N_{int}$ provides an approximation of the RHS function $f$ on its domain, that is $N_{int}(t,x) \approx f(t,x)$.

\begin{figure}[H]
    \centering
    \includegraphics[width = 1\linewidth]{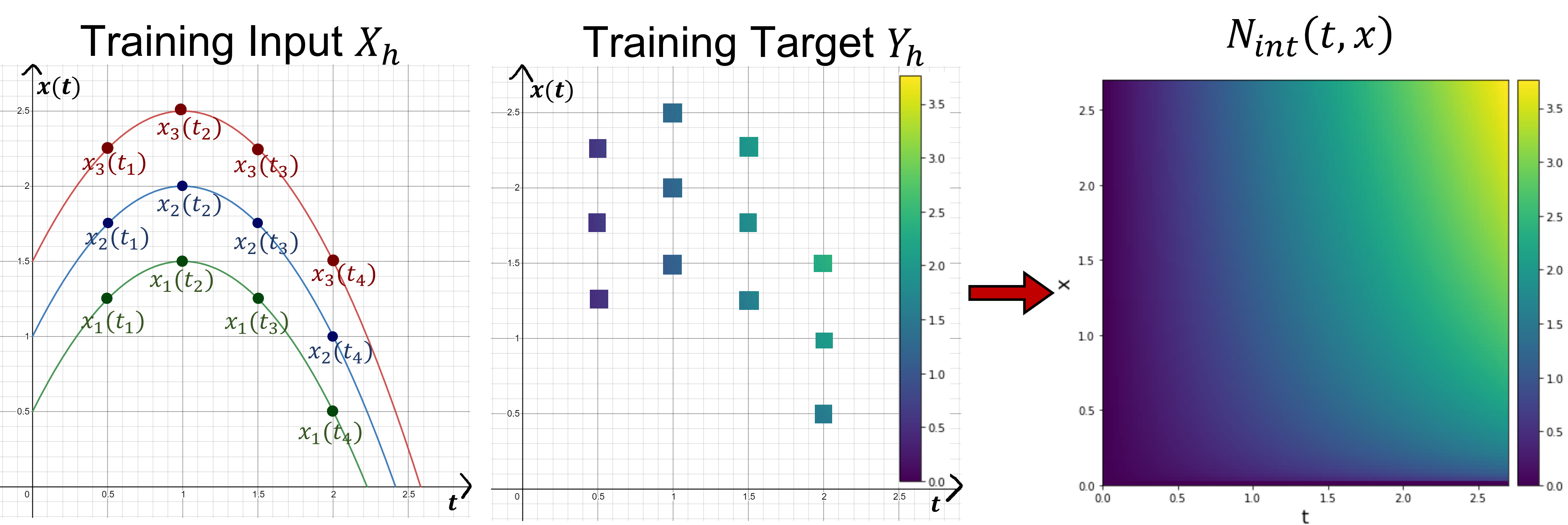}
    \caption{A representation of the data for the interpolation network: the inputs are observations of a time $t$ and of the state vector $x(t)$, the target are discrete approximations of the velocity vector $\dot{x}(t)$  which act as a prior for the values of the unknown function $f(t,x)$. The goal is to reconstruct the function $f$ on its domain using the Lipschitz regularized neural network $N_{int}$ only from the discrete approximations of the velocity vector.}
    \label{fig:data_interp}
\end{figure}

The data used by the interpolation network are couples $(X_h, Y_h), \; h = j + (i-1)M = 1, \dots, KM$, where $X_h$ is the  input and $Y_h$ is the target and $X_h$, $Y_h$ are defined as follows:
\begin{align*}
&X_h = (t_j, \; x_i(t_j)) \in \mathbb{R}^{1+d},\\
&Y_h =  (\dot{x}_i(t_j)) \in \mathbb{R}^{d}.
\end{align*}
 The data is separated into training and testing sets made respectively of 80\% and 20\% of the data. A representation of the data for the interpolation network is provided in Figure \ref{fig:data_interp}.
The interpolation network is feed forward neural network with $L$ layers and Leaky ReLU activation function. We apply the network to each training input $X_h$ and we aim to find the best network parameters to match the corresponding $Y_h$.\\
For $ i = 1, \dots, L$ define the weight matrices $W_{int}^i \in \mathbb{R}^{\;n_i \times n_{i-1}}$ and bias vectors $b_{int}^i \in \mathbb{R}^{n_i}$ where $n_i \in \mathbb{N}, n_0 = 1+d, n_L = d$. Let $\theta_{int} = \{W_{int},b_{int}\}$ be the model parameters.\\
For an input $X_h \in \mathbb{R}^{1+d}$ and parameters $\theta$ we have: 
$$N_{int}(X_h, \theta_{int}) = W_{int}^L(\dots W_{int}^3\sigma (W_{int}^2\sigma ( W_{int}^1X_h +b_{int}^1) +b_{int}^2)\dots) + b_{int}^L \; \in \mathbb{R}^{d}.$$

The loss function minimized to train the interpolation network contains two terms. The first one is the Mean Squared Error (MSE) between the network output and the target data: this forces the network predictions to be close to the observed data. The second term is a a Lipschitz regularization term which forces the Lipschitz constant of the network $N_{int}$ to be small. In contrast with the most common choices of regularization terms found in Machine Learning literature, we don't impose an explicit regularization on the network parameters, but we impose a Lipschitz regularization on the statistical geometric mapping properties of the network. More details about this regularization term can be found in our paper \cite{NEGRINI2021110549}.
Specifically, the loss function has the form:
$$L(\theta_{int}) = \frac{1}{KM}\sum_{h=1}^{KM} \| Y_h - N_{int}(X_h, \theta_{int}) \|^2_2 + \alpha \text{Lip}(N_{int}),$$
where $\| \cdot \|_2$ is the $L^2$ norm, $\alpha >0$ is a regularization parameter and $\text{Lip}(N_{int})$ is the Lipschitz constant of the network $N_{int}$. The predicted approximation of the function $f(t,x)$ is given by the network $N_{int}$ corresponding to $\underset{\theta_{int}}{\mathrm{argmin}}\, L(\theta_{int})$.\\
The Lipschitz constant of the network $N_{int}$, $\text{Lip}(N_{int})$, is computed as:
$$\text{Lip}(N_{int}) = \|\nabla N_{int}\|_{L^{\infty}(\mathbb{R}^{d+1})}$$
where the gradient of the network with respect to the input $X_h$ is computed exactly using \texttt{autograd} \cite{NEURIPS2019_9015}.
We note that controlling the Lipschitz constant of the network $N_{int}$ yields control on the smoothness and rate of change of the approximating function.

In the examples we approximate the Lipschitz constant of the interpolation network using an approach similar to the one exposed in \cite{calliess2015bayesian}: a finite set $S$ of points is selected randomly in the domain of $f$ where the data was generated; then, the Lipschitz norm of the network is estimated as the infinity norm of the gradient of $N_{int}$ evaluated on $S$.
The approximation of the derivative of the network $N_{int}$ with respect to its inputs is computed in Python using \texttt{autograd} \cite{NEURIPS2019_9015}. 
Note that, as empirically shown in \cite{calliess2015bayesian}, the larger the cardinality of the set $S$, the better the approximation of the Lipschitz constant. In our experiments, we set the cardinality of $S$ to be 1000.

\begin{remark}
For ease of notation, in the rest of the paper we will drop the explicit dependence of the networks from their learnable parameters and we will only write $N_j(X_i^j)$ and $N_{int}(X_h)$.
\end{remark}

\section{Synthetic Data and Model Evaluation}\label{data_eval}
In this section we describe the synthetic data used in the experiments and the metrics we use to evaluate the performance of the ensemble architecture.
\subsection{Data Generation}
In the numerical examples we use synthetic data generated in Python:  using the function \texttt{odeint} from the \texttt{scipy} package in Python (\cite{2020SciPy-NMeth}), we solve $\dot{x}(t) = f(t, x(t))$;\; this provides us with approximations of the state vector $x(t)$ for initial conditions  $x_1(0), \dots, x_K(0) \in \mathbb{R}^d$ at time steps $t_1,\dots, t_M$. We perform the experiments in the case of noiseless data, and data with up to $10\%$ of noise. To generate noisy data, we proceed as follows: for each component $x^k(t)$ of the solution $x(t)$ we compute its mean range $M_k$ across trajectories as $$M_k =\frac{1}{K}\left(\sum_{i=1}^K |\max_{j=1,\dots,M} x_i^k(t_j) - \min_{j=1,\dots,M} x_i^k(t_j)|\right).$$ 
Then, the $5\%$ noisy version of $x_i^k(t_j)$ is given by $$\hat{x}_i^k(t_j) = x_i^k(t_j) + n_{ij}M_k, $$ where $n_{ij}$ is a sample from a normal distribution $\mathscr{N}(0,0.05)$ with mean 0 and variance 0.05. In a similar way we add $10\%$ of noise to the data.
\subsection{Model Evaluation}\label{eval}
We use three different metrics to evaluate the performance of the ensemble architecture.
\begin{enumerate}
    \item We use the Mean Squared Error (MSE) on test data which measures the distance of the ensemble prediction from the test data. We note that in a real-world problem the \textit{test error} is the only information accessible to evaluate the performance on the model. We  also report the \textit{generalization gap} obtained with and without Lipschitz regularization in the interpolation network. The generalization gap  measures the ability of the ensemble network to generalize to unseen data (for a more precise description see section \ref{gen_gap}).
    
    \item Since we only use synthetic data, we have access to the true RHS function $f(t,x)$. This allows to compute the relative MSE between the true $f(t,x)$ and the approximation given by the ensemble architecture on arbitrary couples $(t,x)$ in the domain of $f$. We call this error \textit{recovery error}. Note that the error obtained in this way may be different than the one obtained using test data since the test data may be influenced by the noise in the original observations, while here we compare with the true values of the function $f$.
    
    \item Since the neural network ensemble produces a function $N_{int}(t,x)$, it can be used as RHS of a differential equation $\dot{x} = N_{int}(t,x)$. We then solve this differential equation in Python and compute the relative MSE between the solution obtained when using as RHS the ensemble approximation $N_{int}(t,x)$ and when using the true function $f(t,x)$. We call this \textit{error in the solution}.
\end{enumerate}



\subsection{Generalization Gap}\label{gen_gap}
In our previous paper \cite{NEGRINI2021110549} we approximated the RHS of a system of differential equations $\dot{x} = f(t,x)$ using a Lipschitz regularized deep neural network and we empirically demonstrated that adding a Lipschitz regularization term in the loss function improves the ability of the model to generalize to unseen data. The neural network used in our previous work had the same structure as the interpolation network proposed here, but the target data was not generated using an ensemble of neural networks. In fact, in our previous work we first denoised the trajectory data using spline interpolation, then we approximated the velocity vector using the numerical derivative of the splines. One limitation of our previous method was that, when large amounts of noise were present in the data or when splines could not approximate the trajectories correctly, the target data obtained with this process did not provide a reliable approximation of the velocity vector. This, in turn, resulted in poor approximations of the RHS function $f$.

In this work we show that not only the approximation of the true RHS function can be improved when using and ensemble architecture, but also that the Lipschitz regularization term in the interpolation network still improves the generalization properties of the model. We do this by comparing, for a fixed training error, the test error and generalization gap obtained by the ensemble with and without Lipschitz regularization. In the following we will precisely define the generalization gap.

We indicate with $\rho$ the true data distribution, with $\mathscr{D}_k$ the training data distribution and with $\mathscr{D}_{test}$ the discrete distribution of test data. By definition the training data distribution $\mathscr{D}_k$ is a discrete approximation of $\rho$ which converges to $\rho$ as the number of data points $k$ tends to infinity. We write $X \sim \rho$ to indicate that the random variable $X$ has distribution $\rho$.
\medskip\\
The Generalization Gap is defined to be the difference:
$$\mathbb{E}_{X \sim \rho}[\| N_k(X) - Y(X) \|^2_2] - \mathbb{E}_{X \sim \mathscr{D}_k}[\| N_k(X) - Y(X) \|^2_2].$$
Here $N_k$ denotes the optimal function learned after minimizing the loss function $L(\theta)$ on the training data $\mathscr{D}_k$.
While the quantity $\mathbb{E}_{X \sim \mathscr{D}_k}[\| N_k(X) - Y(X) \|^2_2]$ can be explicitly evaluated using the optimal $N_k$ and the training data $\mathscr{D}_k$, the quantity $\mathbb{E}_{X \sim \rho}[\| N_k(X) - Y(X) \|^2_2]$ is unknown since we do not have access to the true data distribution $\rho$. 
In practice, however, the quantity $\mathbb{E}_{X \sim \rho}[\| N_k(X) - Y(X) \|^2_2]$ can be estimated using a test set of data $\mathscr{D}_{\text{test}}$. This is a discrete data set that was not used during the training process, but that faithfully represents the true data density $\rho$, i.e. the discrete distribution $\mathscr{D}_{\text{test}}$ converges, as the number of test data goes to infinity, to the true distribution $\rho$. The optimal network $N_k$ is then evaluated on the test set and the value of  $\mathbb{E}_{X \sim \mathscr{D}_{\text{test}}}[\| N_k(X) - Y(X) \|^2_2]$, is taken as an estimate of $\mathbb{E}_{X \sim \rho}[\| N_k(X) - Y(X) \|^2_2]$.

The estimate of $\mathbb{E}_{X \sim \rho}[\| N_k(X) - Y(X) \|^2_2]$ through $\mathbb{E}_{X \sim \mathscr{D}_{\text{test}}}[\| N_k(X) - Y(X) \|^2_2]$ is more precise the larger is the test data set. More precisely, the Hoeffding inequality (see \cite{abu2012learning}, section 1.3) gives a bound which depends on the number of test data on this approximation: if $m$ is the number of test data, given any $\varepsilon >0$ the Hoeffding inequality states that:
$$ \mathbb{P}(|\mathbb{E}_{X \sim \rho}[\| N_k(X) - Y(X) \|^2_2] - \mathbb{E}_{X \sim \mathscr{D}_{\text{test}}}[\| N_k(X) - Y(X) \|^2_2]|>\varepsilon)\leq 2e^{-2 \varepsilon^2 m}.$$
Justified by this inequality, in our numerical examples we use $\mathbb{E}_{X \sim \mathscr{D}_{\text{test}}}[\| N_k(X) - Y(X) \|^2_2]$ as an estimate of $\mathbb{E}_{X \sim \rho}[\| N_k(X) - Y(X) \|^2_2]$.
\section{Numerical Examples}\label{num_ex}
In this section we propose a few numerical examples of our method and comparisons with other methods for system identification. In the examples we use synthetic data with noise amount up to 10\% for one-dimensional examples, and up to 2\% for higher dimensional examples. In this paper we only propose one and two dimensional examples, but we explicitly notice that, since our method is applied componentwise, it can be used for data of any dimension. Because of the course of dimensionality, however, the higher the dimensionality of the problem and the noise amount, the larger is the amount of data and trainable parameters needed to obtain accurate results. This is the reason why for the two-dimensional examples proposed here we only add up to 2\% of noise in the data.
When using Lipschitz regularization, we considered multiple Lipschitz regularization parameters and selected them using the same heuristic used in \cite{NEGRINI2021110549} Section 4, paragraph 2.

As explained in Section \ref{eval}, we use three different metrics to evaluate the performance of our method.
Specifically, for each example we report the training and testing MSE, the generalization gap and the Estimated Lipschitz constant obtained for Lipschitz regularized and non-regularized ensemble architectures.
Moreover we use the MSE both for the \textit{recovery error} and for the \textit{error in the solution} since this allows to compare such errors with the \textit{test MSE}.\\
Finally, we compare the recovery errors and errors in the solutions obtained by our proposed methods and other methods for system identification. Specifically, we compare our results with our previous method proposed in \cite{NEGRINI2021110549}, with the multistep method proposed in \cite{raissi2018multistep}, with polynomial regression and with the method SINDy proposed in \cite{brunton2016discovering}.

The following examples here are representative of a much larger testing activity in which several different types of right-hand-sides $f(t,x)$, sampling time intervals and initial conditions have been used, leading to comparable experimental results.

\subsection{Empirical Assessment of the Ensemble Algorithm}
In this section we use the three metrics mentioned above to assess the effectiveness of our ensemble algorithm. We also empirically demonstrate that adding a Lipschitz regularization term in the loss function when training the interpolation network improves generalization: this confirms the findings of our previous paper \cite{NEGRINI2021110549}.
\subsubsection{One-dimensional Example}
The first example we propose is the recovery of the ODE 
\begin{equation}\label{one-dim}
    \dot{x}= xe^t +\sin(x)^2 -x
\end{equation}
We generated the data by computing an approximated solution $x(t)$ for the equation (\ref{one-dim}) using the \texttt{odeint} function in Python. We generate solutions for time steps $t$ in the interval [0,0.8] with $\Delta t = 0.04$ and for 500 initial conditions uniformly sampled in the interval [-3,3]. The hyperparameters for our model are selected in each example by cross validation; in this example the interpolation network $N_{int}$ has $L =8$ layers, each layer has 20 neurons, while each network $N_j$ of the target data generator ensemble has $L_j = 3$ layers with 10 neurons each. The target data generator is made of 20 networks.

In Tables \ref{tab:5xet}, \ref{tab:10xet} we report the training MSE, testing MSE, Generalization Gap and estimated Lipschitz constant when 5\% and 10\% of noise is present in the data. We generated these results in a way similar to our previous paper: since our goal here is to compare the performance on test data of the networks with and without regularization, we select the number of epochs during training so as to achieve the same training MSE across all the regularization parameters choices and compare the corresponding Testing errors and Generalization Gaps. We report here only the results obtained for the non-regularized case and for the best regularized one when 5\% and 10\% of noise is present in the data; we already showed in our previous paper that Lipschitz regularization is especially useful in presence of noise, so we omit the noiseless case. We can see from the tables that Lipschitz regularization improves the generalization gap by one order of magnitude for all amounts of noise, that a larger regularization parameter is needed when more noise is present in the data and that, as expected, adding Lipschitz regularization results in a smaller estimated Lipschitz constant. This confirms the findings from our previous paper that Lipschitz regularization improves generalization and avoids overfitting, especially in presence of noise in the data.

\begin{table}[H]
\centering
\begin{tabular}{|c|c|c|c|c|}
\hline
\multicolumn{5}{|c|}{\textit{\begin{tabular}[c]{@{}c@{}}$\dot{x}= xe^t +\sin(x)^2 -x$, 5\% Noise\end{tabular}}}                                                                                                                                                                                                          \\ \hline
\textit{\begin{tabular}[c]{@{}c@{}}Regularization\\ Parameter\end{tabular}} & \textit{\begin{tabular}[c]{@{}c@{}} Training MSE\end{tabular}} & \textit{\begin{tabular}[c]{@{}c@{}} Testing MSE\end{tabular}} & \textit{\begin{tabular}[c]{@{}c@{}} Generalization Gap\end{tabular}}&
\textit{\begin{tabular}[c]{@{}c@{}} Estimated\\ Lipschitz Constant\end{tabular}}\\ \hline
\textit{0}                                                                  & 0.618\%                                                                          & 0.652\%                                                                         & 0.034\%  & 7.09                                                                          \\ \hline
\textit{\textbf{0.004}}                                                     & \textbf{ 0.618\%}                                                                 & \textbf{0.619\%}                                                                & \textbf{0.001\%}       & \textbf{6.33 }                                                           \\ \hline
\end{tabular}
\caption{Test error and Generalization Gap comparison for 5\% noise in the data.}
\label{tab:5xet}
\end{table}

\begin{table}[H]\label{gap10xet}
\centering
\begin{tabular}{|c|c|c|c|c|}
\hline
\multicolumn{5}{|c|}{\textit{\begin{tabular}[c]{@{}c@{}}$\dot{x}= xe^t +\sin(x)^2 -x$, 10\% Noise\end{tabular}}}                                                                                                                                                                                                                      \\ \hline
\textit{\begin{tabular}[c]{@{}c@{}}Regularization\\ Parameter\end{tabular}} & \textit{\begin{tabular}[c]{@{}c@{}} Training MSE\end{tabular}} & \textit{\begin{tabular}[c]{@{}c@{}} Testing MSE\end{tabular}} & \textit{\begin{tabular}[c]{@{}c@{}} Generalization Gap\end{tabular}}&
\textit{\begin{tabular}[c]{@{}c@{}} Estimated \\ Lipschitz Constant\end{tabular}} \\ \hline
\textit{0}                                                                  & 2.01\%                                                                          &2.32\%                                                                         & 0.310\%   & 7.72                                                                           \\ \hline
\textit{\textbf{0.015}}                                                     & \textbf{2.01\%}                                                                 & \textbf{2.03\%}                                                                & \textbf{0.030\%}  & \textbf{6.38}                                                                 \\ \hline
\end{tabular}
\caption{Test error and Generalization Gap comparison for 10\% noise in the data.}
\label{tab:10xet}
\end{table}

In Table \ref{tab:recSolext} we report the error in the recovery for the RHS function $f(t,x) = xe^t +\sin(x)^2 -x$ and the error in the solution of the ODE when using the interpolation network as RHS. We can see that for all amounts of noise in the data, both the reconstruction error and the error in the solution are small, respectively they are less than 0.7\% and 0.04\%.

\begin{table}[H]
\centering
\begin{tabular}{|c|c|}
\hline
\multicolumn{2}{|c|}{\textit{\begin{tabular}[c]{@{}c@{}}Relative MSE in the recovery of the RHS of\\ $\dot{x}=xe^t +\sin(x)^2 -x $\end{tabular}}}                                                                                                                               \\ \hline
\textit{0\% Noise}  & {0.100\%} \\ \hline
\textit{5\% Noise}  & {0.144\%} \\ \hline
\textit{10\% Noise} & {0.663\%} \\ \hline
\end{tabular}
\qquad
\begin{tabular}{|c|c|}
\hline
\multicolumn{2}{|c|}{\textit{\begin{tabular}[c]{@{}c@{}}Relative MSE in the solution of\\ $\dot{x}= xe^t +\sin(x)^2 -x$\end{tabular}}}                                                          \\ \hline
\textit{0\% Noise}  & {0.016\%} \\ \hline
\textit{5\% Noise}  & {0.025\%} \\ \hline
\textit{10\% Noise} & {0.038\%}  \\ \hline
\end{tabular}
\caption{\textbf{Left:} Relative MSE in the recovery of the RHS for up to 10\% of noise. \textbf{Right:} Relative MSE in the solution of the ODE for up to 10\% of noise}
\label{tab:recSolext}
\end{table}

\begin{figure}[H]
 \centering
\includegraphics[width = 3.2in,height=2.2in]{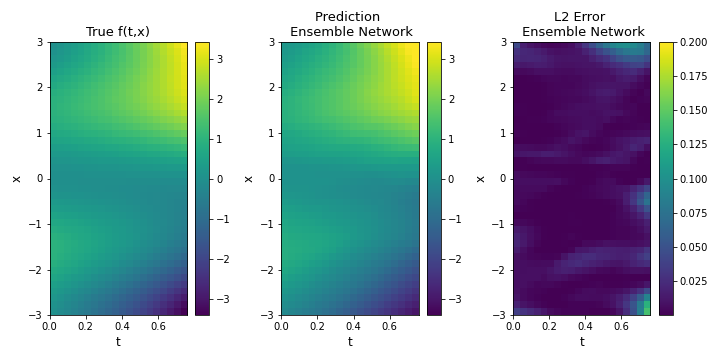}
\includegraphics[width = 2.2in,height=2.2in]{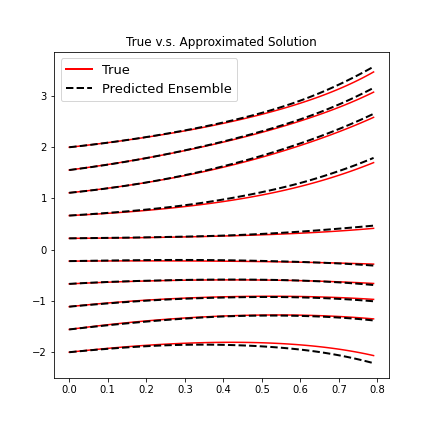}
\caption{ \textbf{Left:} True RHS, Predicted RHS and recovery error for 5\% noise in the data. \textbf{Right:} True and Predicted solution for 5\% noise in the data}
\label{fig:recSolext}
\end{figure}

The left panel of figure \ref{fig:recSolext} shows the true and reconstructed RHS and recovery error on the domain on which the original data was sampled for 5\% of noise in the data. In the error plot a darker color represents a smaller error. We can see that the largest error is attained at the right boundary of the domain: this is due to the fact that by design of our architecture the target data generator only generates target data up to the second-last time step. As a consequence the interpolation network has only access to observations up to the second-last time step and so it is forced to predict the value of the RHS function at the last time step by extrapolation. It is then reasonable that the largest recovery error is attained at the right boundary of the domain. In the right panel of figure \ref{fig:recSolext} we report the true solution (red line) and the solution predicted when using the interpolation network as RHS (dashed black line) for multiple initial condition and for 5\% noise in the data. We notice that the prediction is accurate for all the initial conditions selected, but that it gets worse towards the end of the time interval. This is due to the inaccurate approximation of the RHS at the right boundary of the time interval.

Finally, since the \textit{test error}, the \textit{error in the recovery} and the \textit{error in the solution} are all measured using MSE, it makes sense to compare such homogeneous measurements. The first thing to notice is that the testing errors are larger than the recovery errors. This shows the ability of our network to avoid overfitting and produce reliable approximations of the true RHS even when large amounts of noise are present in the data. In fact, the Test MSE is computed by comparing the value predicted by the network with the value of the corresponding \textit{noisy} observation, while the recovery error is computed by comparing the value predicted by the network with the value of the \textit{true} function $f$. The disparity between the test error and the recovery error then shows that the interpolation network provides results that successfully avoid fitting the noise in the data. The second thing to notice is the disparity between the recovery error and the solution error: specifically the solution error is on average smaller than the recovery error. This is due to the data sampling: when recovering the RHS we reconstruct the function on the full domain, while the original data was only sampled on discrete trajectories; for this reason large errors are attained in the parts of the domain where no training data was available. On the other hand the error in the solution is computed on trajectories which were originally part of the training set, so it is reasonable to expect a smaller error in this case.

\subsubsection{Simple Pendulum}
The second example we propose is the recovery of a simple pendulum equation described by the system of ODEs 
\begin{align*}
    \begin{cases} \dot{x_1} = x_2 \\ \dot{x_2} = -0.5x_1  \end{cases}
\end{align*}
In the notation established above, we let $f = (f_1,f_2)$ with $f_1: = x_2$ and $f_2 := -0.5x_1$.
We generated the data by computing an approximated solution $x(t)$ for the system using the \texttt{odeint} function in Python. We generate solutions for time steps $t$ in the interval [0,0.8] with $\Delta t = 0.04$ and for 1000 initial conditions uniformly sampled in the square $[0,10]\times[0,10]$.  

In this example, the interpolation networks $N_1$, $N_2$ have respectively $L_1 = 10 $ and $L_2 = 10$ layers and each layer has 20 neurons, while each network $N_j$ of the target data generator ensemble has $L_j = 5$ layers with 60 neurons each. The target data generator is made of 20 networks. 

Because of the curse of dimensionality in this case we need more data and more trainable parameters than in the previous example and we only add up to 2\% of noise.

In Tables \ref{tab:pend1N}, \ref{tab:pend2N} we report the training MSE, testing MSE, Generalization Gap and estimated Lipschitz constant when 1\% and 2\% of noise is present in the data. Similarly to what observed in the one-dimensional case, in all cases Lipschitz regularization results in a smaller generalization gap, and estimated Lipschitz constant and that when more noise is present in the data a stronger regularization (that is a larger regularization parameter) is needed.

\begin{table}[H]
\centering
\begin{tabular}{|c|c|c|c|c|}
\hline
\multicolumn{5}{|c|}{\textit{Simple Pendulum, 1\% Noise}}                                                                                                                                                                                                                                                                                                                                          \\ \hline
\multicolumn{5}{|c|}{\textit{Component 1}}                                                                                                                                                                                                                                                                                                                                                        \\ \hline
\textit{\begin{tabular}[c]{@{}c@{}}Regularization\\ Parameter\end{tabular}} & \textit{\begin{tabular}[c]{@{}c@{}} Training MSE\end{tabular}} & \textit{\begin{tabular}[c]{@{}c@{}}Testing MSE\end{tabular}} & \textit{\begin{tabular}[c]{@{}c@{}} Generalization Gap\end{tabular}} & \textit{\begin{tabular}[c]{@{}c@{}}Estimated \\ Lipschitz Constant\end{tabular}} \\ \hline
\textit{0}                                                                  & 0.547\%                                                                  & 0.628\%                                                                & 0.081\%                                                                        &  2.62                                                                         \\ \hline
\textit{\textbf{0.002}}                                                    & \textbf{0.547\%}                                                         & \textbf{0.584\%}                                                       & \textbf{0.037\%}                                                               & \textbf{1.46}                                                                  \\ \hline
\multicolumn{5}{|c|}{\textit{Component 2}}                                                                                                                                                                                                                                                                                                                                                        \\ \hline
\textit{\begin{tabular}[c]{@{}c@{}}Regularization\\ Parameter\end{tabular}} & \textit{\begin{tabular}[c]{@{}c@{}} Training MSE\end{tabular}} & \textit{\begin{tabular}[c]{@{}c@{}} Testing MSE\end{tabular}} & \textit{\begin{tabular}[c]{@{}c@{}} Generalization Gap\end{tabular}} & \textit{\begin{tabular}[c]{@{}c@{}} Estimated \\ Lipschitz Constant\end{tabular}} \\ \hline
\textit{0}                                                                  & 0.408\%                                                                  & 0.452\%                                                                & 0.044\%                                                                        & 3.77                                                                           \\ \hline
\textit{\textbf{0.001}}                                                    & \textbf{0.408\%}                                                         & \textbf{0.412\%}                                                       & \textbf{0.004\%}                                                               & \textbf{0.95}                                                                  \\ \hline
\end{tabular}
\caption{Test error and Generalization Gap comparison for 1\% noise in the data, both components.}
\label{tab:pend1N}
\end{table}

\begin{table}[H]
\centering
\begin{tabular}{|c|c|c|c|c|}
\hline
\multicolumn{5}{|c|}{\textit{Simple Pendulum, 2\% Noise}}                                                                                                                                                                                                                                                                                                                                          \\ \hline
\multicolumn{5}{|c|}{\textit{Component 1}}                                                                                                                                                                                                                                                                                                                                                        \\ \hline
\textit{\begin{tabular}[c]{@{}c@{}}Regularization\\ Parameter\end{tabular}} & \textit{\begin{tabular}[c]{@{}c@{}} Training MSE\end{tabular}} & \textit{\begin{tabular}[c]{@{}c@{}} Testing MSE\end{tabular}} & \textit{\begin{tabular}[c]{@{}c@{}} Generalization Gap\end{tabular}} & \textit{\begin{tabular}[c]{@{}c@{}} Estimated \\ Lipschitz Constant\end{tabular}} \\ \hline
\textit{0}                                                                  & 0.3366\%                                                                  & 0.3775\%                                                                & 0.0409\%                                                                        & 3.02                                                                          \\ \hline
\textit{\textbf{0.008}}                                                    & 0.3366\textbf{\%}                                                         & \textbf{0.3374\%}                                                       & \textbf{0.0008\%}                                                               & \textbf{1.02}                                                                  \\ \hline
\multicolumn{5}{|c|}{\textit{Component 2}}                                                                                                                                                                                                                                                                                                                                                        \\ \hline
\textit{\begin{tabular}[c]{@{}c@{}}Regularization\\ Parameter\end{tabular}} & \textit{\begin{tabular}[c]{@{}c@{}} Training MSE\end{tabular}} & \textit{\begin{tabular}[c]{@{}c@{}} Testing MSE\end{tabular}} & \textit{\begin{tabular}[c]{@{}c@{}} Generalization Gap\end{tabular}} & \textit{\begin{tabular}[c]{@{}c@{}} Estimated \\ Lipschitz Constant\end{tabular}} \\ \hline
\textit{0}                                                                  & 0.3811\%                                                                  & 0.397\%                                                                & 0.016\%                                                                        & 1.11                                                                          \\ \hline
\textit{\textbf{0.006}}                                                    & \textbf{0.3811\%}                                                         & \textbf{0.3814\%}                                                       & \textbf{0.0003\%}                                                               & \textbf{0.84}                                                                  \\ \hline
\end{tabular}
\caption{Test error and Generalization Gap comparison for 2\% noise in the data, both components.}
\label{tab:pend2N}
\end{table}

In Table \ref{tab:recSolPend} we report recovery error for the RHS functions $f_1(t,x) = x_2$ and $f_2(t,x) = -0.5x_1$ and the error in the solution of the ODE when using the interpolation network to approximate $f_1$ and $f_2$. Also in this case we attain a good accuracy both in the recovery of the RHS and in the approximation of the solution with errors for both components respectively less than 0.7\% and 0.07\%. When the noise increases from 1\% to 2\% we observe an increase of one order of magnitude in both the errors in the recovery and in the solution. This shows that when larger amounts of noise are present in the data more data is needed in order to reconstruct accurately the true RHS function.
\begin{table}[H]
\centering
\begin{tabular}{|c|c|c|}
\hline
\multicolumn{3}{|c|}{\textit{\begin{tabular}[c]{@{}c@{}}Relative MSE in the recovery of the RHS of the \\ Simple Pendulum \end{tabular}}}                                                                                                                 \\ \hline
\textit{\textbf{}}  & \textit{\begin{tabular}[c]{@{}c@{}}  Component 1\end{tabular}} & \textit{\begin{tabular}[c]{@{}c@{}} Component 2\end{tabular}} \\ \hline
\textit{0\% Noise}  &  0.0131\%                                                                                   & 0.0186\%                                                                                  \\ \hline
\textit{1\% Noise}  & 0.0468\%                                                                                   & 0.0597\%                                                                                  \\ \hline
\textit{2\% Noise} &0.533\%                                                                                  & 0.645\%                                                                                 \\ \hline
\end{tabular}
\qquad
\begin{tabular}{|c|c|c|}
\hline
\multicolumn{3}{|c|}{\textit{\begin{tabular}[c]{@{}c@{}}Relative MSE in the solution of \\ Simple Pendulum \end{tabular}}}                                                                                                                 \\ \hline
\textit{\textbf{}}  & \textit{\begin{tabular}[c]{@{}c@{}}  Component 1\end{tabular}} & \textit{\begin{tabular}[c]{@{}c@{}} Component 2\end{tabular}} \\ \hline
\textit{0\% Noise}  & 0.002\%                                                                                   & 0.002\%                                                                                  \\ \hline
\textit{1\% Noise}  & 0.004\%                                                                                   & 0.004\%                                                                                  \\ \hline
\textit{2\% Noise} & 0.061\%                                                                                  & 0.051\%                                                                                 \\ \hline
\end{tabular}
\caption{\textbf{Left:} Relative MSE in the recovery of the RHS for up to 2\% of noise. \textbf{Right:} Relative MSE in the solution of the system of ODEs for up to 2\% of noise.}
\label{tab:recSolPend}
\end{table}

\begin{figure}[H]
 \centering
\includegraphics[width = 5in,height=2in]{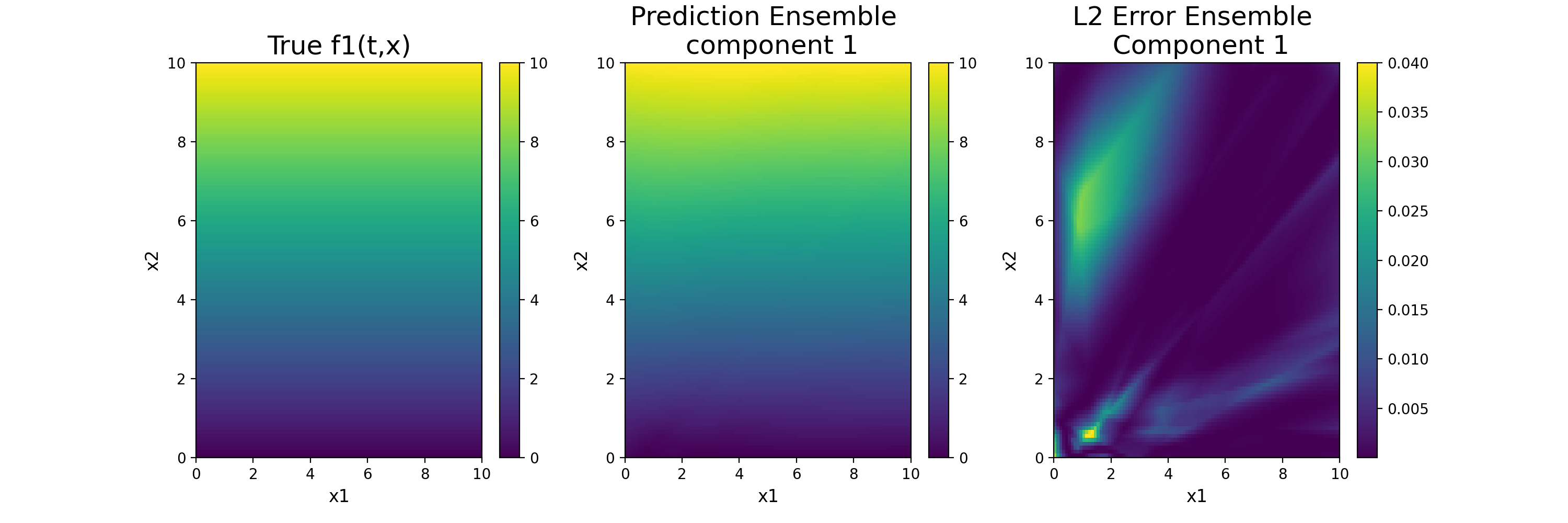} 
\includegraphics[width = 5in,height=2in]{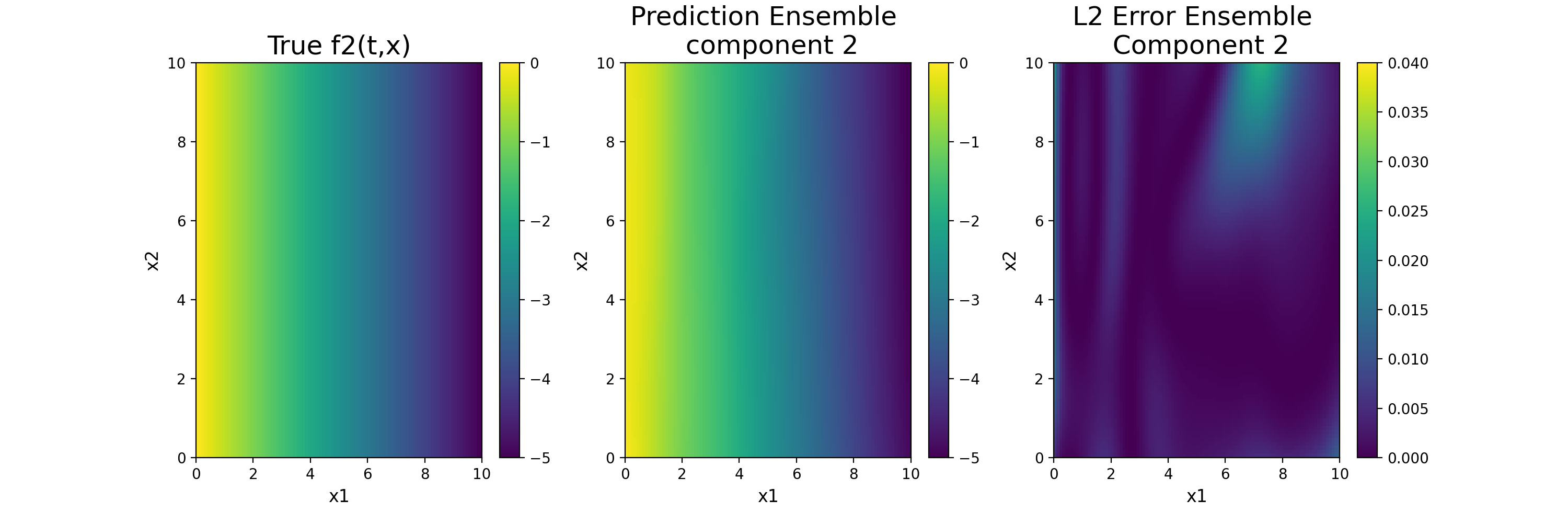}
\caption{Reconstruction of the RHS function $f_1, f_2$ when 1\% of noise is present in the data. \textbf{Left:} True RHS functions $f_1,f_2$. \textbf{Center: } Reconstructed RHS functions. \textbf{Right:} Error for the two components.}
\label{fig:RHSPend}
\end{figure}

\begin{figure}[H]
 \centering
\includegraphics[width = 5in,height=2in]{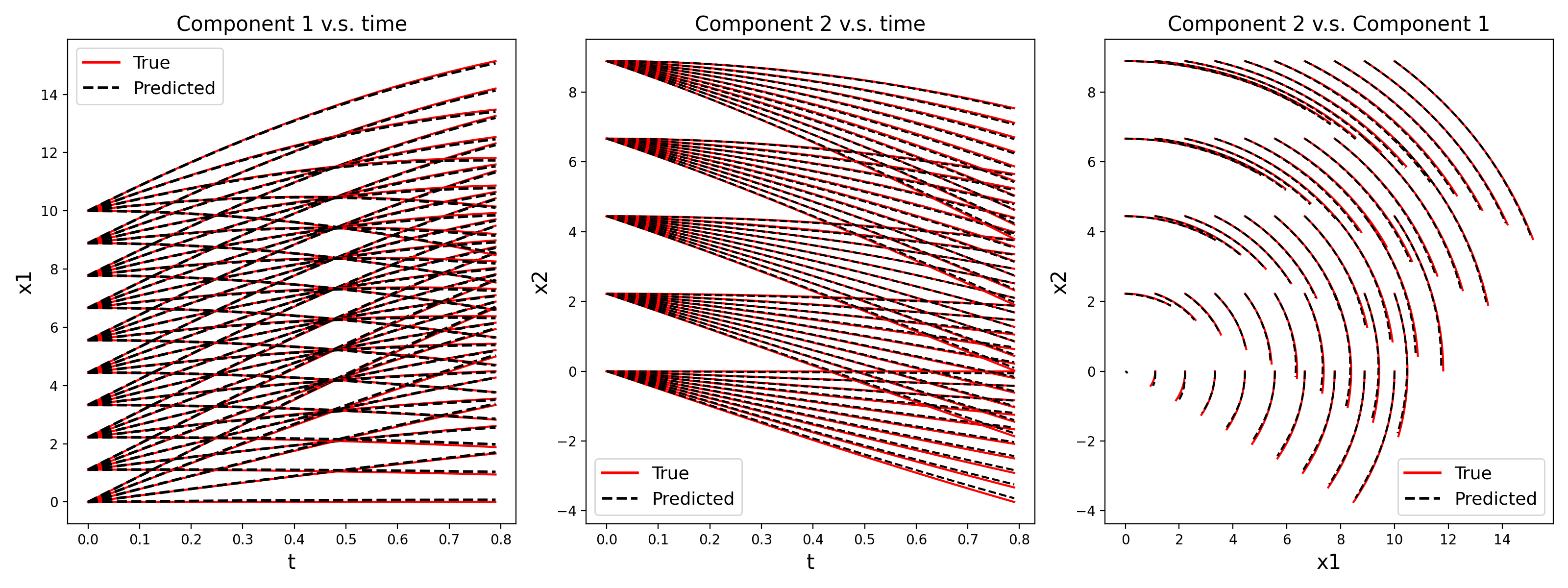} 
\caption{True and Predicted solutions when $1\%$ Noise is present in the data. \textbf{Left:} component 1 v.s. time. \textbf{Center:} component 2 v.s. time. \textbf{Right:} component 2 v.s. component 1.}
\label{fig:solPend}
\end{figure}

In Figure \ref{fig:RHSPend} we show the reconstructed RHS functions $f_1, f_2$ when 1\% noise is present in the data. We note from these plots that the error in the RHS recovery is small across the whole domain for both components showing the ability of our ensemble method to prevent overfitting. In Figure \ref{fig:solPend} we show the true and predicted solutions $x_1, x_2$ for multiple initial conditions obtained when using the network approximations of $f_1$ and $f_2$ as RHS functions in the ODE solver. We observe that the true and predicted solutions are nearly indistinguishable from each other with the largest disparity between the two happening for large values of $t$.


Finally, as in the one-dimensional case we observe that the test errors are larger than the errors in the RHS recovery for all amounts of noise, showing that indeed when noise is present in the data our method is able to avoid overfitting providing reliable approximations of the true RHS function. Again as before the discrepancy between the error in the RHS approximation and in the solution is due to the different data sapling in the two scenarios (see previous example for a more precise explanation). We also note that the error in the RHS reconstruction and in the solution are closely related and influence each other: if the approximation of the RHS is poor, then also the approximation of the solution will be of poor quality.

\subsection{Comparison with the Splines Method}\label{previousPap}
As explained in Section \ref{Intro}, in our previous paper \cite{NEGRINI2021110549} (\textit{splines method}) we approximated the RHS of a system of differential equations $\dot{x} = f(t,x)$ using a Lipschitz regularized deep neural network. The architecture used in our previous work is the same as the interpolation network proposed here, however, the target data, approximations of the velocity vector, is generated differently. Instead of using an ensemble of neural networks, in our previous work we first denoise the trajectory data using spline interpolation, then we approximate the velocity vector using the numerical derivative of the splines. We showed that this approach is very effective when the trajectories can be correctly approximated by splines. However, when this is not true, for example if trajectories are non-smooth in $t$ or if large amounts of noise is present in the data, the target data obtained using splines derivatives did not provide a reliable approximation of the velocity vector. This, in turn, resulted in poor approximations of the RHS function.

We explicitly notice that the only difference between the splines method and the ensemble method is the way we preprocess the data: in the splines method we use splines to produce reliable target data, while in the ensemble method we use the target data generator. For both methods we then use a Lipschitz regularized neural network to generate the approximation for the RHS function. To fairly compare the two methods we use the same number of trainable parameters for the splines network and for the interpolation network.

In this section we show examples for which our previous method fails at providing a good approximation of the RHS function, but for which our ensemble method succeeds.

\subsubsection{Non-smooth Right-hand Side}\label{nsRHS}
We propose is the recovery through Lipschitz approximation of
\begin{equation}\label{sign}
    \dot{x}= \text{sign}(t-0.1)
\end{equation}
Both the ensemble and the splines method aim at learning a Lipschitz approximation of the right-hand side function. The spline method is based on the notion of classical solution and it is doomed to fail in such a non-smooth setting. In contrast, the ensemble method is based on weak notion of solution using integration as in formula (\ref{weakEq}).

We generated the data by computing an approximated solution $x(t)$ for the equation (\ref{sign}) using the \texttt{odeint} function in Python. We generate solutions for time steps $t$ in the interval [0,0.2] with $\Delta t = 0.02$ and for 500 initial conditions uniformly sampled in the interval [-0.1,0.1] for noise amounts up to 2\%. We only use up to 2\% of noise since, as explained in our previous paper, the splines model can only provide reliable target data for small noise amounts.The hyperparameters for the models in this example are as follows: each network $N_j$ of the target data generator ensemble has $L_j = 3$ layers with 10 neurons each, the interpolation network and the network used in the splines method both have $L =4$ layers, each layer has 30 neurons.  The target data generator is made of 10 networks.

\begin{figure}[H]
 \centering
 \includegraphics[width = 4.8in,height=1.3in]{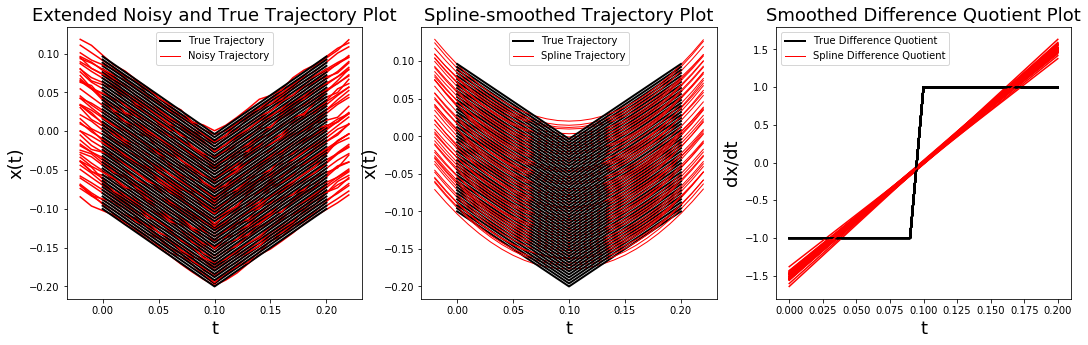} 
\caption{\textbf{Left:} true (black) and noisy trajectories (red). \textbf{Center:} true trajectories (black) and spline approximation (red) of the noisy trajectories. \textbf{Right:} True derivative (black) and spline derivative (red). Since the trajectories are non-smooth in $t$ the trajectory and derivative approximations obtained using splines are poor.}
\label{fig:nonsmooth}
\end{figure}

\begin{figure}[H]
 \centering
\includegraphics[width = 3.5in,height=1.5in]{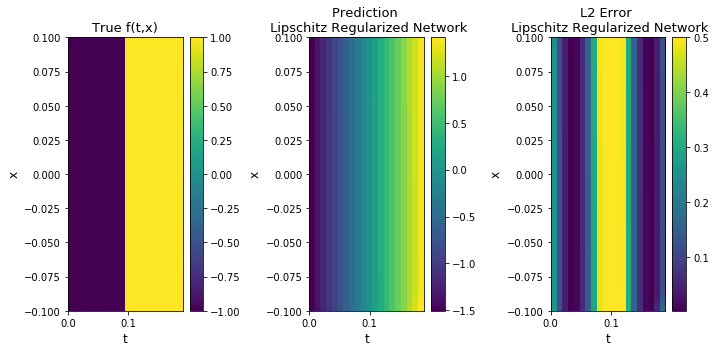} 
\includegraphics[width = 3.5in,height=1.5in]{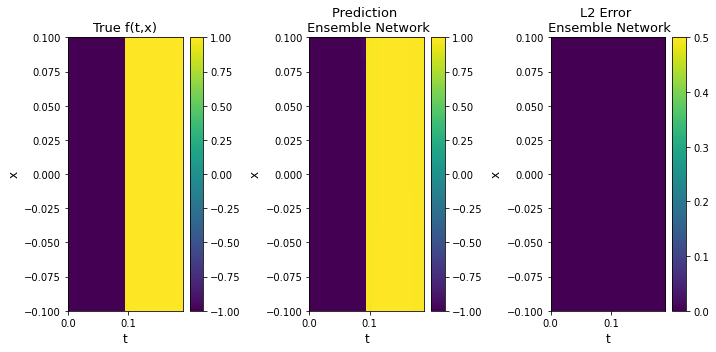}
\caption{\textbf{Top row:} Spline method. \textbf{Bottom row:} Ensemble method. From left to right: True RHS, Reconstructed RHS and Error in the reconstruction when 1\% of noise is present in the data.}
\label{fig:rec_nonsmooth}
\end{figure}

Figure \ref{fig:nonsmooth} shows how the low quality spline approximation of the trajectory data (Center) obtained in the pre-processing stage results in a completely wrong velocity approximation (Right). Note that, while it is clear from this plot that the derivative approximation obtained using the splines is wrong since we have access to the true difference quotients (black line), when using real world data we have no access to the true trajectories or to the true derivatives so it may not be as easy to detect when the splines produce low quality target data. On the other hand, our new ensemble method is completely data-driven and overcomes this approximation difficulty through the use of the target data generator ensemble and the intergal notion of solution introduced in equation (\ref{weakEq}).
Because of the low quality of the target data obtained by spline interpolation, we can see from Table \ref{tab:nonsmooth} that the error in the recovery of the RHS for the method that uses splines is around 12\% for all amounts of noise, while when using our ensemble method it is lower than 0.005\%. The superior performance of the ensemble method over the spline method for this example can also be seen from Figure \ref{fig:rec_nonsmooth}. In this figure from left to right we represent the true, reconstructed RHS and the error in the reconstruction for the spline based method (top row) and for the ensemble method (bottom row) when 1\% of noise is present in the data. We can see from the figure that the spline method in this case is not even able to find the general form of the RHS function correctly because of the bad quality of the target data. On the contrary, our proposed ensemble method, being completely data driven and based on a weak notion of solution, is able to accurately reconstruct RHS functions like $\text{sign}(t-0.1)$ that are non-smooth in $t$.
\begin{table}[H]
\centering
\begin{tabular}{|c|c|l|c|}
\hline
\multicolumn{4}{|c|}{\textit{\begin{tabular}[c]{@{}c@{}}Relative MSE in the recovery of the RHS of\\ $\dot{x}=\text{sign}(t)$\end{tabular}}}                                                                                                                 \\ \hline

\textit{}          & \multicolumn{2}{c|}{\textit{\begin{tabular}[c]{@{}c@{}} Ensemble\end{tabular}}} & \textit{\begin{tabular}[c]{@{}c@{}}  Splines\end{tabular}} \\ \hline
\textit{1\% Noise} & \multicolumn{2}{c|}{0.002\%}                                                                    & 12.5\%                                                                                  \\ \hline
\textit{2\% Noise} & \multicolumn{2}{c|}{0.004\%}                                                                    & 12.9\%                                                                                  \\ \hline
\end{tabular}
\caption{Relative MSE in the recovery of the RHS for up to 2\% of noise for the Ensemble and Splines methods.}
\label{tab:nonsmooth}
\end{table}

\subsubsection{Highly Oscillatory Right-hand Side}
We compare our ensemble method with the splines method for an equation with highly oscillatory RHS function. We propose is the recovery of 
\begin{equation}\label{oscillEq}
    \dot{x}= cos(50t)x
\end{equation}
We generate solutions of \ref{oscillEq} for time steps $t$ in the interval [0,0.2] with $\Delta t = 0.02$ and for 500 initial conditions uniformly sampled in the interval [-0.1,0.1] for noise amounts up to 2\%. The hyperparameters for the models are as follows: each network $N_j$ of the target data generator ensemble has $L_j = 3$ layers with 10 neurons each, the interpolation network and the network used in the splines method both have $L =4$ layers, each layer has 30 neurons.  The target data generator is made of 10 networks.

\begin{figure}[H]
 \centering
 \includegraphics[width = 5in,height=1.5in]{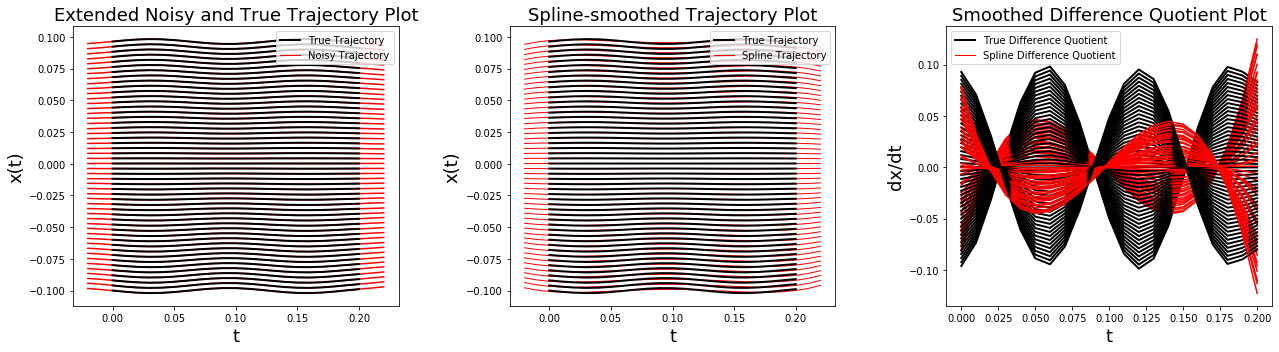} 
\caption{\textbf{Left:} true (black) and noisy trajectories (red). \textbf{Center:} true trajectories (black) and spline approximation (red) of the noisy trajectories. \textbf{Right:} True derivative (black) and spline derivative (red). }
\label{fig:oscill}
\end{figure}

In this example we see that even if in the pre-processing stage the spline approximation of the trajectory data seems accurate from the central panel in Figure \ref{fig:oscill}, the derivative approximation is not because of its highly oscillatory nature, as can be seen in the right panel of Figure \ref{fig:oscill}.

\begin{figure}[H]
 \centering
\includegraphics[width = 3.5in,height=1.5in]{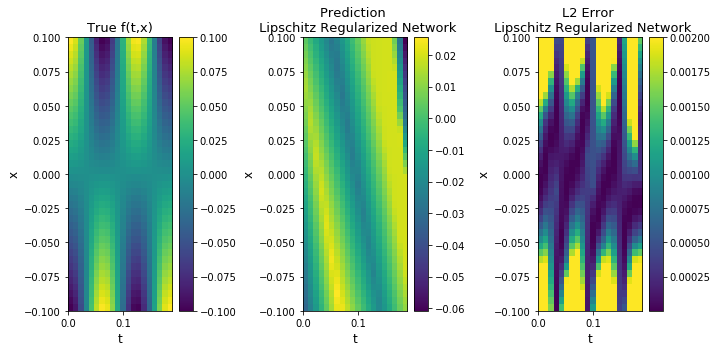} 
\includegraphics[width = 3.5in,height=1.5in]{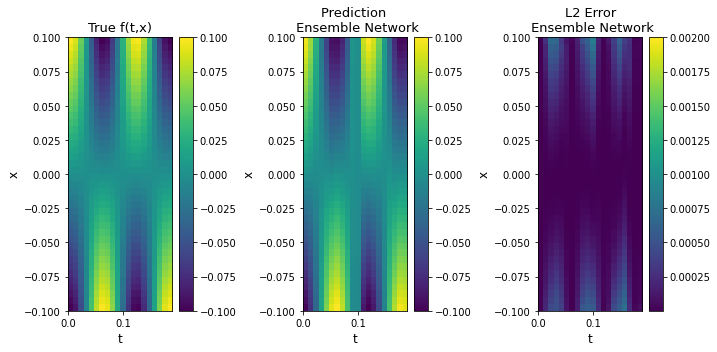}
\caption{\textbf{Top row:} Spline method. \textbf{Bottom row:} Ensemble method. From left to right: True RHS, Reconstructed RHS and Error in the reconstruction when 1\% of noise is present in the data.}
\label{fig:rec_oscill}
\end{figure}

The bad quality target data for the spline model results in errors in the RHS reconstruction of 0.5\% and 0.6\% respectively for 1\% and 2\% of noise in the data. The ensemble model instead provides more accurate reconstructions with errors in the recovery of 0.04\% and 0.05\% respectively (see Table \ref{tab:oscill}). Finally Figure \ref{fig:rec_oscill} represents, from left to right, the true, reconstructed RHS and the error in the reconstruction for the spline method (top row) and for the ensemble method (bottom row) when 1\% of noise is present in the data. We can see that while the ensemble method is able to reconstruct correctly the RHS function, the spline method is not even able to correctly identify the oscillatory nature of the RHS function.

\begin{table}[H]
\centering
\begin{tabular}{|c|c|l|c|}
\hline
\multicolumn{4}{|c|}{\textit{\begin{tabular}[c]{@{}c@{}}Relative MSE in the recovery of the RHS of\\ $\dot{x}= cos(50t)x$\end{tabular}}}
\\ \hline
\textit{}          & \multicolumn{2}{c|}{\textit{\begin{tabular}[c]{@{}c@{}} Ensemble\end{tabular}}} & \textit{\begin{tabular}[c]{@{}c@{}}  Splines\end{tabular}} \\ \hline
\textit{1\% Noise} & \multicolumn{2}{c|}{0.042\%}                                                                    & 0.505\%                                                                                  \\ \hline
\textit{2\% Noise} & \multicolumn{2}{c|}{0.054\%}                                                                    & 0.599\%                                                                                  \\ \hline
\end{tabular}\caption{Relative MSE in the recovery of the RHS for up to 2\% of noise for the Ensemble and Splines methods. }
\label{tab:oscill}
\end{table}

\subsection{Comparison with other methods}\label{comparisonR}
We compare our method with the methods proposed in \cite{raissi2018multistep} and in \cite{brunton2016discovering}. For completeness we also provide a comparison with the splines method \cite{NEGRINI2021110549}. The method proposed in \cite{raissi2018multistep}, (\textit{multistep method}), is similar to ours: the authors place a neural network prior on the RHS function $f$, express the differential equation in integral form and use a multistep method to predict the solution at each successive time steps. The main difference with our method is that we use an ensemble of neural networks as a prior for $f$ instead of a single neural network $f$. We compare the ensemble and the multistep methods using Euler integral form for the equation and the same number of learnable parameters for the multistep method and the interpolation network. Similar results can be obtained when using multistep methods like Adams–Bashforth or Adams–Moulton to represent the equation in integral form.\\
The method proposed in \cite{brunton2016discovering}, which we refer to as \textit{SINDy}, is based on a sparsity-promoting technique: sparse regression is used to determine, from a dictionary of basis functions, the terms in the dynamic governing equations which most accurately represent the data.\\
Finally, we compare with the splines method described before.\\
We report here the relative error obtained by the different methods in the approximation of $f$ as well as in the solution of the ODE. The test error and generalization gaps for the ensemble model were also computed for these examples and confirmed our previous findings: as before we noticed an improvement in the generalization gap when Lipschitz regularization was added.

\subsubsection{Non-Linear, Autonomous Right-hand-side}\label{SINDy_comp}
We generated the data by computing approximated solutions of 
\begin{equation}
    \dot{x}= cos(3x) +x^3 -x
\end{equation} for time steps $t$ in the interval [0,1] with $\Delta t = 0.04$ and for 500 initial conditions uniformly sampled in the interval $[-0.7,0.9]$. The interpolation network $N$ has $L= 8$ layers, each layer has 30 neurons, while each network $N_j$ of the target data generator ensemble has $L_j = 3$ layers with 20 neurons each. The target data generator is made of 25 networks.

We compare the results obtained by the ensemble, spline , multistep methods, a polynomial regression with degree 20 and SINDy. SINDy allows the user to define custom dictionaries of functions to approximate an unknown differential equations from data. In this case we used a custom library of functions containing polynomials up to degree 10 as well as other elementary functions such as $\sin(x),\, \cos(x),\, e^x,\, \ln(x)$.
\smallskip\\
In Table \ref{tab:reccomp3}  we report the relative MSE in the recovery of the RHS function $f= cos(3x) +x^3 -x$ for up to 10\% of noise when using our ensemble method, the splines method, the multistep method, a polynomial regression with degree 20 and SINDy with a custom library. We notice that when no noise is present in the data, so that overfitting is not a concern, SINDy outperforms all the other methods, followed by the polynomial regression. On the contrary, when noise is present in the data our ensemble method gives the best results. For example, when 5\% noise is present in the data our ensemble method obtains an error of 0.096\%  which is smaller than the errors obtained by all the other methods by one order of magnitude or more. This shows that the ensemble method is able to overcome the sensitivity to noise. 

 \begin{table}[H]
 \centering
\begin{tabular}{|c|c|c|c|c|c|}
\hline
\multicolumn{6}{|c|}{\textit{\begin{tabular}[c]{@{}c@{}}Relative MSE in the recovery of the RHS of\\ $\dot{x}= cos(3x) +x^3 -x$\end{tabular}}}                                                                                                         \\ \hline
\textit{}           & \textit{\begin{tabular}[c]{@{}c@{}} Ensemble (Ours)\end{tabular}} & \textit{\begin{tabular}[c]{@{}c@{}} Splines\end{tabular}} & \textit{\begin{tabular}[c]{@{}c@{}} Multistep (Euler)\end{tabular}} & \textit{\begin{tabular}[c]{@{}c@{}} Polynomial Regression\\ degree 20\end{tabular}} & \textit{\begin{tabular}[c]{@{}c@{}} SINDy\\ custom library 
\end{tabular}} \\ \hline
\textit{0\% Noise}  & 0.0505\%                                                                        & 0.214\%                                                                              & 0.116\%                                                                   & 6.3e-05\%                                                                                         & { \textbf{5.7e-05\%}}                                              \\ \hline
\textit{5\% Noise}  & { \textbf{0.0957\%}}                                        & 0.585\%                                                                              & 1.20\%                                                                    & 3.33\%                                                                                            & 0.762\%                                                                                \\ \hline
\textit{10\% Noise} & { \textbf{0.520\%}}                                         & 1.90\%                                                                               & 3.51\%                                                                    & 17.0\%                                                                                            & 3.36\%                                                                                 \\ \hline
\end{tabular}
\caption{Relative MSE in the recovery of the RHS for up to 10\% of noise for ensemble, splines and multistep methods, polynomial regression with degree 20, SINDy with custom library. }
\label{tab:reccomp3}
\end{table}

\begin{figure}[H]
\includegraphics[height=1in, width =\linewidth]{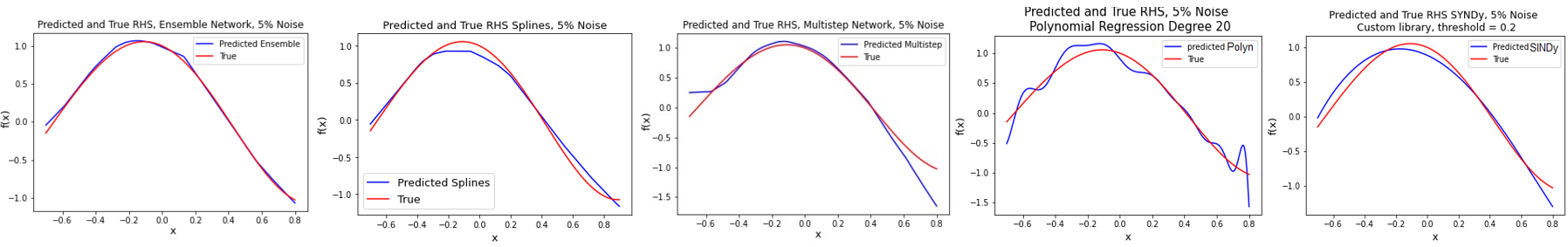}
\caption{From left to right, true and recovered RHS for $5\%$ noise in the data obtained by Ensemble Method (Ours), splines method, Multistep Method, Polynomial Regression with degree 20, SINDy with custom library.}
\label{fig:reccomp3}
\end{figure}

In Figure \ref{fig:reccomp3} we report the true (red line) and recovered RHS function (blue line) when 5\% of noise is present in the data when using the ensemble, spline, multistep methods, polynomial regression with degree 20 and SINDy. This figure confirms the findings shown in the previous table: the ensemble network is able to reconstruct the true RHS most accurately showing that our method is robust to noise. From the table above we notice that, for noisy data, the worst accuracy was always attained by the polynomial regression. In this case, even if a 20 degree polynomial has 100 times less parameters than our neural network, increasing the degree of the polynomial increased the error in the recovery. From this figure we can clearly see why that happens: the polynomial regression with degree 20 is already overfitting the noisy data and the largest errors are attained at the boundaries of the domain where the polynomial is highly oscillatory. The other three methods are able to provide approximations that capture the general form of the true RHS function, but only our ensemble method is able to provide an accurate approximation even at the boundary of the domain.

Finally in Table \ref{tab:solcomp3} we report the relative MSE in the solution of the ODE when using as RHS of the ODE solver the approximation given by the ensemble, spline and multistep models, the polynomial regression and SINDy. When no noise is present in the data, SINDy and polynomial regression provide the best results. When noise is present in the data again our multistep method gives the best results since it is able to overcome the sensitivity to noise.

\begin{table}[H]
\begin{tabular}{|c|c|c|c|c|c|}
\hline
\multicolumn{6}{|c|}{\textit{\begin{tabular}[c]{@{}c@{}}Relative MSE in the solution of\\ $\dot{x}= cos(3x) +x^3 -x$\end{tabular}}}                                                                                                         \\ \hline
\textit{}           & \textit{\begin{tabular}[c]{@{}c@{}} Ensemble (Ours)\end{tabular}} & \textit{\begin{tabular}[c]{@{}c@{}} Splines\end{tabular}} & \textit{\begin{tabular}[c]{@{}c@{}} Multistep (Euler)\end{tabular}} & \textit{\begin{tabular}[c]{@{}c@{}} Polynomial Regression\\ degree 20\end{tabular}} & \textit{\begin{tabular}[c]{@{}c@{}} SINDy\\ custom library 
\end{tabular}} \\ \hline
\textit{0\% Noise}  & 0.00313\% & 0.0289\%                                                                          & 0.00342\%                                                                             &0.00033\%                                                                                             & \textbf{1e-05\%}                                                                         \\ \hline
\textit{5\% Noise}  & \textbf{0.0123\%} &0.0637\%                                                                 & 0.0366\%                                                                             & 0.312\%                                                                                              & 0.965\%                                                                                  \\ \hline
\textit{10\% Noise} & \textbf{0.142\%}    &0.954\%                                                              & 0.251\%                                                                             & 3.11\%                                                                                              & 0.359\%                                                                                  \\ \hline
\end{tabular}
\caption{Relative MSE in the solution of the ODE for up to 10\% of noise for ensemble, splines and multistep methods, polynomial regression with degree 20, SINDy with custom library. }
\label{tab:solcomp3}
\end{table}

\subsubsection{Non-linear, Non-autonomous Right-hand Side}
For this example we generated the data by computing approximated solutions of
\begin{equation}
    \dot{x}= t\cos(x) +t^2x
\end{equation}
for time steps $t$ in the interval [0,1.2] with $\Delta t = 0.04$ and for 500 initial conditions uniformly sampled in the interval $[-2,2]$.  The interpolation network $N$ has $L= 8$ layers, each layer has 30 neurons, while each network $N_j$ of the target data generator ensemble has $L_j = 3$ layers with 20 neurons each. The target data generator is made of 30 networks.

In Table \ref{tab:recSolcomp1} we report the relative MSE in the recovery of the RHS function $f= t\cos(x) +t^2x$ (left table) and in the solution of the ODE (right table) for up to 10\% of noise when using the ensemble, multistep and splines methods. From the tables we can see that the ensemble method outperforms the other two algorithms both in the recovery of the RHS and in the approximation of the ODE solution. For example, when 10\% of noise is present in the data the recovery error for the ensemble method is 0.8\% while for the multistep and spline methods it is respectively less than 1.75\% and 1.1\%. Similarly, our ensemble method attains the best ODE solution accuracy for all amounts of noise.

\begin{table}[H]
\centering
\resizebox{\textwidth}{!}{
\begin{tabular}{|c|c|c|c|}
\hline
\multicolumn{4}{|c|}{\textit{\begin{tabular}[c]{@{}c@{}}Relative MSE in the recovery of the RHS of\\ $\dot{x}= t\cos(x) +t^2x$\end{tabular}}}                                                                                                                               \\ \hline
\textit{\textbf{}}  & \textit{\begin{tabular}[c]{@{}c@{}} Ensemble (Ours)\end{tabular}} & \textit{\begin{tabular}[c]{@{}c@{}} Multistep (Euler)\end{tabular}}& \textit{\begin{tabular}[c]{@{}c@{}} Splines\end{tabular}} \\ \hline
\textit{0\% Noise}  & \textbf{0.074\%}                                                                  & 0.281\% & 0.116\%                                                                   \\ \hline
\textit{5\% Noise}  & \textbf{0.147\%}                                                                  & 0.906\%   & 0.440 \%                                                                \\ \hline
\textit{10\% Noise} & \textbf{0.807\%}                                                                  & 1.758\%  &  1.10 \%                                                                \\ \hline
\end{tabular}
\qquad
\begin{tabular}{|c|c|c|c|}
\hline
\multicolumn{4}{|c|}{\textit{\begin{tabular}[c]{@{}c@{}}Relative MSE in the solution of\\ $\dot{x}= t\cos(x) +t^2x$\end{tabular}}}                                                          \\ \hline
\textit{}           & \textit{\begin{tabular}[c]{@{}c@{}} Ensemble (Ours)\end{tabular}} & \textit{\begin{tabular}[c]{@{}c@{}} Multistep (Euler)\end{tabular}}& \textit{\begin{tabular}[c]{@{}c@{}} Splines \end{tabular}} \\ \hline
\textit{0\% Noise}  & \textbf{0.023\%}                                                                  &  0.048\%     & 0.028\%                                                                        \\ \hline
\textit{5\% Noise}  & \textbf{0.041\%}                                                                  & 0.491\%  & 0.163\%                                                                          \\ \hline
\textit{10\% Noise} & \textbf{0.216\%}                                                                  & 0.856\%   &     0.592\%                                                                     \\ \hline
\end{tabular}}
\caption{\textbf{Left:} Relative MSE in the recovery of the RHS for up to 10\% of noise for ensemble, multistep and splines method.  \textbf{Right:} Relative MSE in the solution of the system of ODEs for up to 10\% of noise for ensemble, multistep and splines method.}
\label{tab:recSolcomp1}
\end{table}

\begin{figure}[H]
 \centering
\includegraphics[width = 3.8in,height=3.2in]{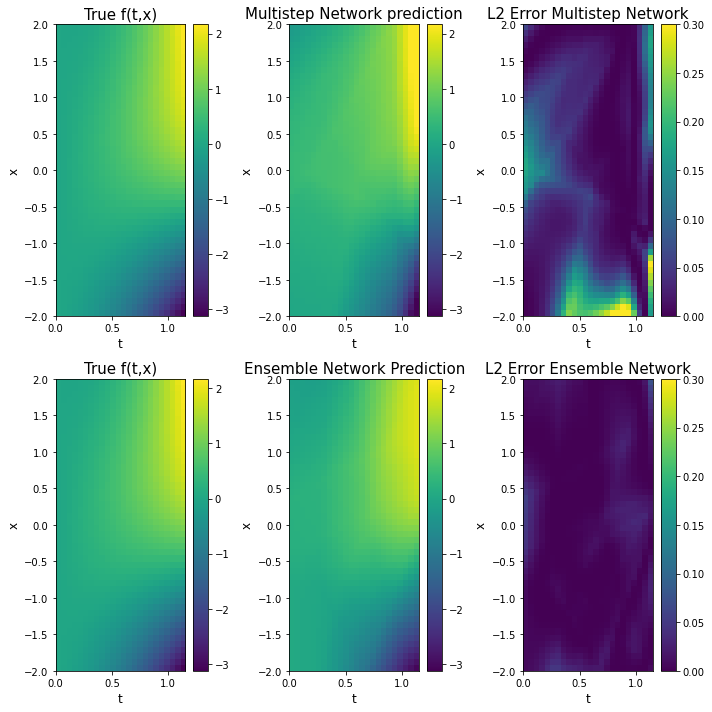}
\caption{\textbf{Top row:} Multistep method. \textbf{Bottom row:} Ensemble method. From left to right: True RHS, Reconstructed RHS and Error in the reconstruction when 5\% of noise is present in the data.}
\label{fig:reccomp1}
\end{figure}

Finally in Figure \ref{fig:reccomp1} we report the true and reconstructed RHS and the error in the reconstruction when 5\% of noise is present in the data, for the multistep method (top row) and for our ensemble method (bottom row). The error plots, where darker color represents a smaller error, show that our esemble method attains a smaller recovery error than the multistep method in the whole domain.

We notice explicitly that, while the results above show that our method is more accurate than those in \cite{raissi2018multistep} and \cite{NEGRINI2021110549}, it is computationally more expensive. In fact in order to generate the target data for the interpolation network, we need to train as many networks as the number of time instants at which we observe the data. On the contrary, for the multistep and spline methods only one network is trained, making these methods faster to train and less computationally expensive than ours.

\section{Discussion}\label{discussion}
In the previous section we proposed multiple numerical examples of our ensemble method. We used one and two dimensional synthetic data with up to 10\% of noise, but since our model is applied componentwise, it can be used for data of any dimension. To evaluate the performance of our method we used three different metrics: the test MSE, the recovery error and the error in the solution. For a precise description of these metrics see Section \ref{eval}.\\
\indent
Our first goal was to compare the performance on test data of the interpolation network with and without Lipschitz regularization and see if the findings of our previous paper still applied in this case where we use an ensemble of neural networks to generate the target data instead of splines. The examples show that indeed Lipschitz regularization improves the generalization ability of the network as well as its ability to overcome sensitivity to noise and to avoid overfitting. Specifically, in all of the examples we observed an average improvement of one order of magnitude in the generalization gap confirming the findings of our previous paper \cite{NEGRINI2021110549} and of \cite{oberman2018lipschitz}.\\
\indent
Next, we studied the recovery error and the error in the solution. We observed that, for all amounts of noise in the data, our ensemble method is able to accurately reconstruct the RHS function on the domain on which the data was sampled, with the largest errors being attained at the right boundary of the time domain where no target data was available. In all of the examples we observed that, for a fixed noise amount, the test error was larger than the recovery error. This shows the ability of our network to avoid overfitting. In fact, the Test MSE measures the difference between the network prediction and the \textit{noisy} observations, while the recovery error measures the difference between the network prediction and the \textit{true} function $f$. A smaller recovery error then shows that the interpolation network provides results that closely fit the true function and successfully avoid fitting the noise in the data.\\
\indent
We also compared the true solution of the equation with the solution obtained when using the approximated RHS in the ODE solver. Also in this case we obtain accurate results in the solution reconstruction with errors that increase close to the end point of the time interval. In all of the examples we observed that the error in the solution is on average smaller than the recovery error. This is due to the data sampling: the recovery of the RHS function is done on the full domain, even if the original data was only sampled on discrete trajectories; for this reason large errors are attained in the parts of the domain where no training data was available. On the other hand the error in the solution is computed only on trajectories which were originally part of the training set, so we obtain smaller errors in this case.\\
\indent
Finally, in Sections \ref{previousPap}, \ref{comparisonR} we compared our ensemble algorithm with the method proposed in our previous paper \cite{NEGRINI2021110549} (splines method), with the method proposed in \cite{raissi2018multistep} (multistep method), with polynomial regression and with SINDy \cite{brunton2016discovering}. In all of the examples proposed our ensemble method is the one that provides the best recovery errors and errors in the solution. Specifically, we show that using the ensemble method we are able to reconstruct RHS functions that our previous paper could not reconstruct correctly, such as functions that are non-smooth in $t$ or with highly oscillatory terms. This is due to the fact that our ensemble method is completely data-driven and based on a weak notion of solution using integration.  Our ensemble method outperforms the multistep method, polynomial regression and SINDy especially when noise is present in the data. We note however, that while our proposed method is more effective in providing accurate RHS approximations, it is more computationally expensive than the other methods we compared to. In fact in order to generate the target data for the interpolation network, we need to train as many networks as the number of time instants at which we observe the data. This can be very expensive especially if the user is interested in reconstructing equations on long time intervals. On the contrary, for the multistep and spline methods only one network is trained, while for polynomial regression and SINDy only a loss minimization is needed to obtain the results, making these methods faster to train than ours. We conclude that the proposed ensemble method is the best choice, compared to the other methods described here, if the goal is to obtain very accurate reconstructions even in presence of noise and the computational cost is not a concern.
 
\section{Conclusion}\label{conclusion}
In this paper we use a Lipschitz regularized ensemble of neural networks to learn governing equations from data. There are two main differences between our method and other neural network system identification methods in the literature. First, we add a Lipschitz regularization term to our loss function to force the Lipschitz constant of the interpolation network to be small. This regularization results in a smoother approximating function and better generalization properties when compared with non-regularized models, especially in presence of noise. These results are in line with the theoretical work of Calder and Oberman \cite{oberman2018lipschitz} and with the empirical findings of our previous paper \cite{NEGRINI2021110549}.  Second, we use an ensemble of neural networks instead of a single neural network for the reconstruction. It has been shown in \cite{hansen1990neural} that the generalization ability of a neural network architecture can be improved through ensembling, but while this technique has been applied in the past for multiple problems (see for instance \cite{shimshoni1998classification}, \cite{huang2000pose}, \cite{zhou2002lung}), to our knowledge this is the first time that Lipschitz regularization is added to the ensemble to overcome the sensitivity to noise in a system identification problem.\\
\indent
The results shown in the examples, which are representative of a larger testing activity with several different types of right-hand sides $f(x,t)$, show multiple strengths of our method:
\begin{itemize}
    \item In all of the examples when noise is present in the data, the Lipschitz regularization term in the loss function results in an improvement of generalization gap of one order of magnitude, when compared to the non-regularized architecture.
    \item The ensemble architecture is robust to noise and is able to avoid overfitting even when large amounts of noise are present in the data (up to 10\%). The ability of the ensemble to avoid overfitting is numerically confirmed by the fact that test errors are larger than recovery errors in all of the examples. In fact, while the test error measures the distance of the ensemble prediction from the noisy RHS data, the recovery error measures the distance to the true RHS data, so that the disparity between test and recovery errors means that the ensemble is able to avoid fitting the noise in the data. This robustness to noise is especially an advantage over methods that do not use ensembling such as \cite{raissi2018multistep}, as can be seen from the examples in Section \ref{comparisonR}.
    \item The ensemble architecture is completely data-driven and it is based on weak notion of solution using integration (see formula \ref{weakEq}). For this reason, it can be used to reconstruct non-smooth RHS functions (see Example \ref{nsRHS}). This is especially an advantage over models that rely on the notion of classical solution like the Splines Method \cite{NEGRINI2021110549}.
    \item Since neural networks are universal approximators, we do not need any prior knowledge on the ODE system, in contrast with sparse regression approaches in which a library of candidate functions has to be defined. As shown in Section \ref{SINDy_comp}, direct comparison with polynomial regression and SINDy shows that our model is a better fit when learning from noisy data. 
    \item Since our method is applied componentwise, it can be used to identify systems of any dimension, which makes it a valuable approach for high-dimensional real-world problems. However, because of the curse of dimensionality, the higher the problem dimension, the larger the amount of data and trainable parameters needed to obtain accurate predictions.
\end{itemize}
We explicitly note that, while our ensemble model is able to reconstruct the RHS function $f$ with high accuracy even for very noisy data, it is computationally more expensive than the other methods we compared with (splines and multistep methods, polynomial regression and SINDy). This is because of the ensemble nature of the algorithm: in order to generate the target data for the interpolation network, we need to train as many networks as the number of time instants at which we observe the data. Consequently, this algorithm is a good choice for applications where high reconstruction accuracy is need but where the computational cost is not a concern.

Future research directions include applying our method to real-world data and generalize it to learn partial differential equations. In contrast with the most common choices of regularization terms found in machine learning literature, in this work we impose a regularization on the network statistical geometric mapping properties, instead of on its parameters. The Lipschitz regularization results, however, in an implicit constraint on the network parameters since the minimization of the loss function is done with respect to such parameters. An interesting future direction is to theoretically study the Lipschitz regularization term, how it relates to the size of the weights of the network, in line with Bartlett work about generalization \cite{bartlett1997valid}, and express it as an explicit constraint on the network learnable parameters. In this way one could avoid approximating the Lipschitz constant of the network numerically. This would decrease considerably the computational cost of the algorithm which, as explained, is one limitation of the proposed method.

\section{Acknowledgements}
Luca Capogna is partially supported by NSF DMS 1955992 and Simons Collaboration Grant for Mathematicians 585688.\\
Giovanna Citti is partially supported by the EU Horizon 2020 project GHAIA,  MCSA RISE project GA No 777822.\\
Results in this paper were obtained in part using a high-performance computing system acquired through NSF MRI grant DMS-1337943 to WPI.\\

\bigskip
\bibliographystyle{plain}
\small
\bibliography{bibib}

\end{document}